\documentclass[10pt,journal,compsoc]{IEEEtran}

\usepackage{url}
\usepackage{times}
\usepackage{epsfig}
\usepackage{graphicx}
\usepackage{amsmath}
\usepackage{amssymb}
\usepackage{multirow}
\usepackage{subfigure}
\usepackage{url}
\usepackage{color}
\usepackage[dvipsnames]{xcolor}
\usepackage{soul}
\usepackage[normalem]{ulem}
\usepackage{enumitem}
\setlist[itemize]{noitemsep, topsep=0pt,left=0pt}

\usepackage{graphicx}
\usepackage{amsmath}
\usepackage{amssymb}
\usepackage{booktabs}
\usepackage{multirow}
\usepackage{makecell}
\usepackage{subfloat}
\usepackage{balance} 
\usepackage{soul}
\usepackage[accsupp]{axessibility}
\usepackage[autostyle]{csquotes}
\usepackage[pagebackref,breaklinks,colorlinks]{hyperref}
\usepackage[capitalize]{cleveref}
\crefname{section}{Sec.}{Secs.}
\Crefname{section}{Section}{Sections}
\Crefname{table}{Table}{Tables}
\crefname{table}{Tab.}{Tabs.}

\usepackage{algorithm}
\usepackage{algorithmic}

\ifCLASSOPTIONcompsoc
  \usepackage[nocompress]{cite}
\else
  \usepackage{cite}
\fi

\ifCLASSINFOpdf

\else

\fi

\begin{document}
\title{Adaptive Fine-Grained Predicates Learning for Scene Graph Generation} 
\author{Xinyu~Lyu,
	Lianli~Gao,
	Pengpeng~Zeng,
	Heng~Tao~Shen~\IEEEmembership{Fellow,~IEEE}, 
	and Jingkuan~Song
	\thanks{Corresponding Author: Lianli Gao} 
}

\markboth{Journal of \LaTeX\ Class Files,~Vol.~14, No.~8, August~2015}%
{Shell \MakeLowercase{\textit{et al.}}: Bare Demo of IEEEtran.cls for Computer Society Journals}

\IEEEtitleabstractindextext{%
\begin{abstract}
The performance of current Scene Graph Generation (SGG) models is severely hampered by hard-to-distinguish predicates, e.g., ``woman-\underline{on}/\underline{standing on}/\underline{walking on}-beach''. As general SGG models tend to predict head predicates and re-balancing strategies prefer tail categories, none of them can appropriately handle hard-to-distinguish predicates. To tackle this issue, inspired by fine-grained image classification, which focuses on differentiating hard-to-distinguish objects, we propose an \textbf{Adaptive Fine-Grained Predicates Learning (FGPL-A)} which aims at differentiating hard-to-distinguish predicates for SGG. First, we introduce an \textbf{Adaptive Predicate Lattice (PL-A)} to figure out hard-to-distinguish predicates, which adaptively explores predicate correlations in keeping with model's dynamic learning pace. Practically, PL-A is initialized from SGG dataset, and gets refined by exploring model’s predictions of current mini-batch. Utilizing PL-A, we propose an \textbf{Adaptive Category Discriminating Loss (CDL-A)} and an \textbf{Adaptive Entity Discriminating Loss (EDL-A)}, which progressively regularize model’s discriminating process with fine-grained supervision concerning model's dynamic learning status, ensuring balanced and efficient learning process. Extensive experimental results show that our proposed model-agnostic strategy significantly boosts performance of benchmark models on VG-SGG and GQA-SGG datasets by up to \textbf{175\% and 76\% on Mean Recall@100}, achieving new state-of-the-art performance. Moreover, experiments on Sentence-to-Graph Retrieval and Image Captioning tasks further demonstrate practicability of our method.

\end{abstract}

\begin{IEEEkeywords}
Scene Graph Generation, Visual Relationship, Fine-Grained Learning, Adaptive Learning.
\end{IEEEkeywords}}

\maketitle

\IEEEdisplaynontitleabstractindextext

\IEEEpeerreviewmaketitle

\IEEEraisesectionheading{\section{Introduction}\label{sec:intro}}
\IEEEPARstart{S}{cene} graph generation plays a vital role in visual understanding, which intends to detect instances together with their relationships. By ultimately representing image contents in a graph structure, scene graph generation serves as a powerful means to bridge the gap between visual scenes and human languages, benefiting several visual-understanding tasks, such as image retrieval~\cite{image_retrival}, image captioning~\cite{cap1,cap2}, and visual question answering~\cite{ssg:vctree,qa1,qa3,qa4}. 

\begin{figure}[htbp]
\begin{center}
\includegraphics[width=0.47\textwidth]{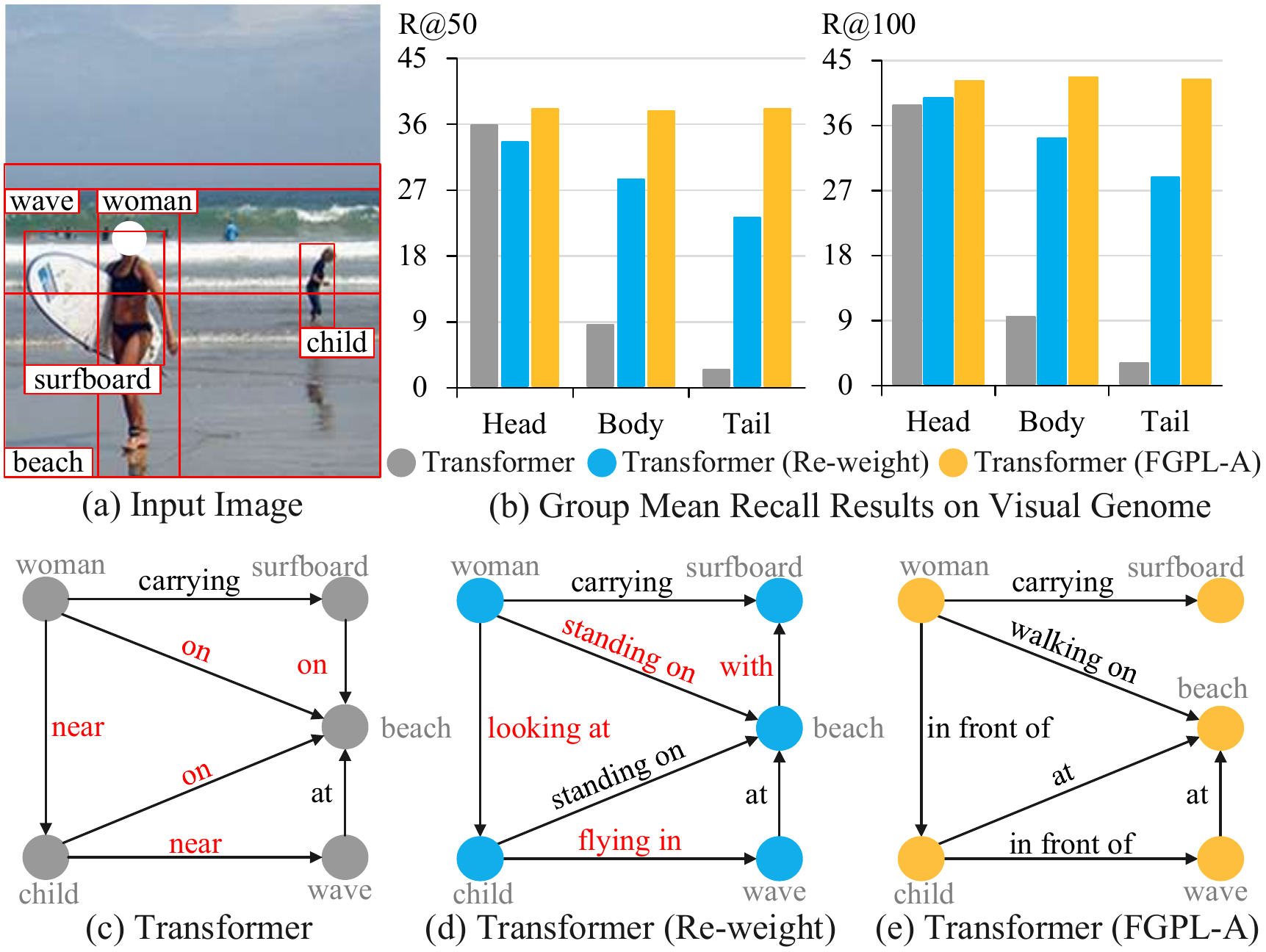}
\caption{\textbf{The illustration of handling hard-to-distinguish predicates for SGG models.} (b) Transformer (FGPL-A) outperforms both Transformer and Transformer (Re-weight) on Group Mean Recall with a balanced discrimination among predicates with different frequencies. (c) Transformer~\cite{networks:transformer,ssg:benchmark} is prone to predict head predicates. (d) Transformer (Re-weight) prefers tail categories. (e) Transformer (FGPL-A) can appropriately handle hard-to-distinguish predicates, e.g., ``woman-on/standing on/\textbf{walking on}-beach'' or ``woman-near/looking at/\textbf{in front of}-child''.}
\label{fig:abstract}
\end{center}
\end{figure}
Prior works~\cite{imp,motifs,ssg:vctree,gps,vrr,energy,icme_chen} have devoted great efforts to exploring representation learning for scene graph generation, but the biased prediction issue is still challenging because of the long-tailed distribution of predicates in SGG datasets. Trained with severely skewed class distributions, general SGG models are prone to predict head predicates, as results of Transformer~\cite{networks:transformer,ssg:benchmark} shown in Fig.~\ref{fig:abstract}(c). Recent works~\cite{cogtree,prior:iccv,bgnn,icme_zheng} have exploited re-balancing methods to solve the biased prediction problem for scene graph generation, making predicates distribution balanced or the learning process smooth. As demonstrated in Fig.~\ref{fig:abstract}(b), Transformer (Re-weight) achieves a more balanced performance than Transformer. However, relying on the class distribution, existing re-balancing strategies prefer predicates from tail categories while being hampered by some hard-to-distinguish predicates.
In fine-grained image classification task, hard-to-distinguish object classes include species of birds, flowers, or animals; and the makes or models of vehicles.
Similarly, typical hard-to-distinguish predicates in SGG include types of human-object-interactions (e.g., ``on'', ``standing on'' or ``walking on''), spatial (e.g., ``in front of'', ``behind'' or ``near'') or possessive relationships (e.g., ``part of'', ``attached to'' or ``covering''). And they cannot be well addressed by either general SGG models or re-balancing methods.
For instance, as shown in Fig.~\ref{fig:abstract}(d), Transformer (Re-weight) misclassifies \textit{``woman-\underline{in front of}-beach''} as \textit{``woman-\underline{looking at}-child''} in terms of visual correlations between ``in front of'' and ``watching'' in this scenario.


The origin of the issue lies in the fact that differentiating among hard-to-distinguish predicates requires exploring their correlations first. 
As an inherent characteristic of predicates, the predicate correlations reflect the difficulty of distinction between predicate pairs, i.e., if two predicates are hard-to-distinguish or not for SGG models.
Different from fine-grained image classification task where hard-to-distinguish object classes correlation are predefined, the hard-to-distinguish predicates correlation in SGG task is unknown.
Hence, without exploration of predicate correlations, the existing re-balancing SGG methods~\cite{scenegraph:gcl,bgnn,dt2} cannot adaptively adjust discriminating process in accordance with the difficulty of distinction, resulting in an inefficient learning process.
Particularly, roughly adjusting discriminating process based on the distribution prior, the dataset-based weights in re-balancing SGG methods may over-emphasize the tail predicates and meanwhile over-suppress the head classes. 
Consequently, the over-adjustment of optimization tends to make model's learning process less effective or even misleading.
Ultimately, as claimed in~\cite{prob:overconfident,overconfident}, the inefficient learning process may cause over-confidence on tail predicates with under-represented head classes, which degrades model's discriminatory power among hard-to-distinguish predicates and thereby deteriorates SGG model's performance on generating fine-grained scene graphs.
For example, in Fig.~\ref{fig:abstract}(d), ``in front of'' is misclassified to ``flying in'' simply because ``flying in'' is a tail predicate with less observations in SGG datasets.

\begin{figure}[t]
\begin{center}
\includegraphics[width=0.4\textwidth]{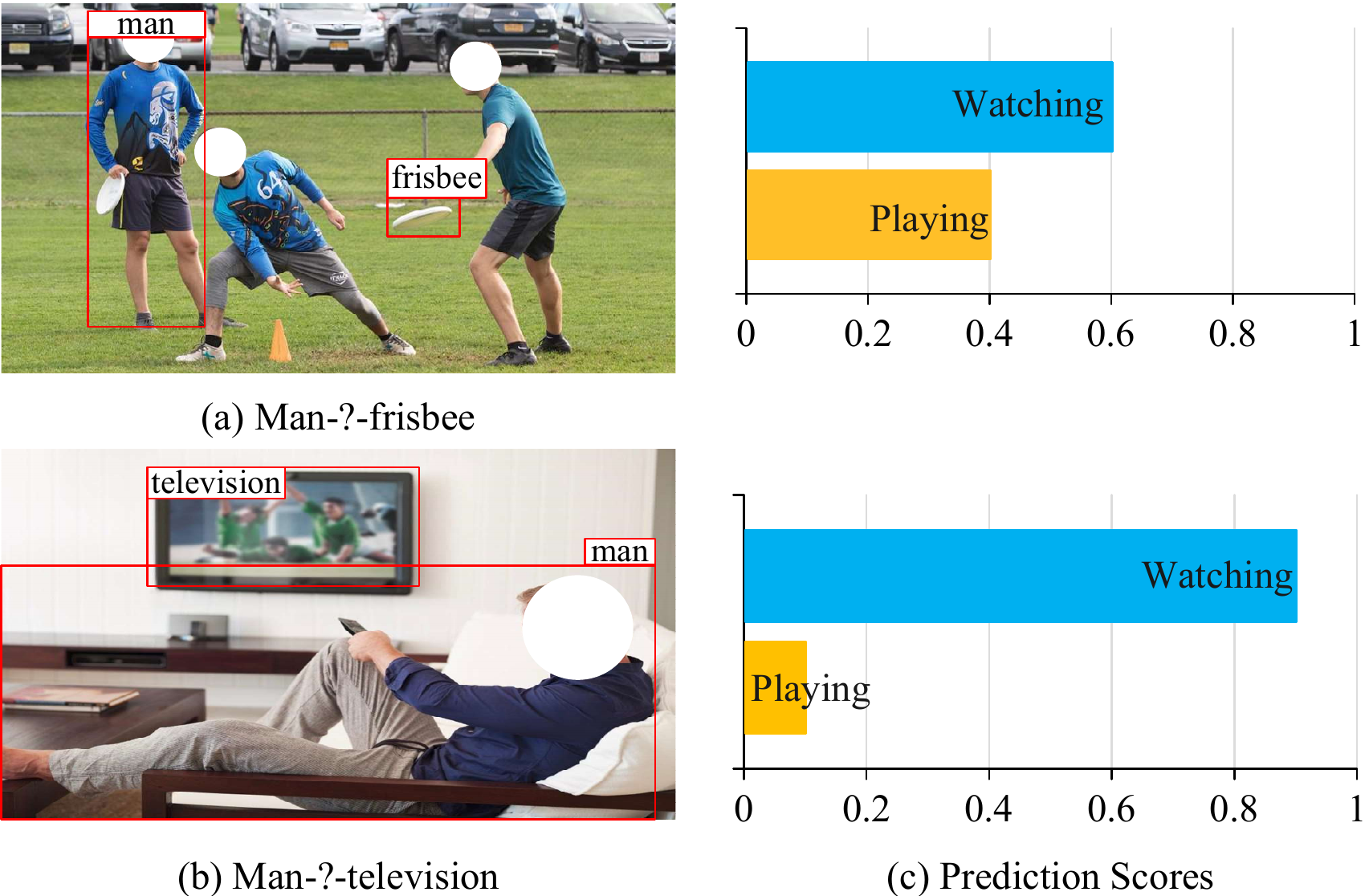}
\caption{\textbf{The illustration of predicate correlations concerning contexts.} The predicate correlations between ``watching'' and ``playing'' vary with contexts. Especially, ``watching/playing'' is weakly correlated or distinguishable in (b), while they are strongly correlated or hard-to-distinguish in (a).}
\label{fig:problem}
\end{center}
\end{figure}

To acquire comprehensive predicate correlations, we consider the contextual information for predicate pairs, since correlations between a pair of predicates may dramatically vary with contexts as stated in~\cite{nlp}. 
Particularly, contexts are regarded as visual or semantic information of predicates' objects and subjects in scene graph generation. 
Take predicate correlations analysis between ``watching'' and ``playing'' as an example. ``Watching/playing'' is weakly correlated or distinguishable in Fig.~\ref{fig:problem}(b), while they are strongly correlated or hard-to-distinguish for SGG models in Fig.~\ref{fig:problem}(a).
Except for the contextual information, the predicate correlations may gradually change during model's learning process.
Revealing whether the predicates are hard-to-distinguish or not, the predicate correlations reflects model's learning status over predicate classes, which varies with its dynamic learning pace in training iterations.
For instance, the predicates pair ``on'' and ``standing on'' is hard-to-distinguish for SGG models at first, and then gradually becomes recognizable during training.
Hence, the predicate correlations need progressive refinement in keeping with model's dynamic learning status for the comprehensive understanding.

Inspired by the above observations, we propose an Adaptive Fine-Grained Predicates Learning (FGPL-A) framework for SGG, which dynamically figures out and discriminates among hard-to-distinguish predicates in keeping with model's dynamic learning pace.
We first introduce an Adaptive Predicate Lattice (PL-A) to help SGG models understand ubiquitous predicate correlations concerning the underlying contextual information and model's dynamic learning status. 
In practice, PL-A is initialized by investigating biased predictions of SGG datasets from general SGG models.
Then, we devise a Batch-Refinement (BR) regime to iteratively refine the PL-A by exploring model's ongoing predictions of the current mini-batch.
Furthermore, by exploring the predicate correlations of PL-A, we devise an Adaptive Category Discriminating Loss (CDL-A) and an Adaptive Entity Discriminating Loss (EDL-A), which both differentiate among hard-to-distinguish predicates while maintaining learned discriminatory power over recognizable ones.
In particular, the Adaptive Category Discriminating Loss (CDL-A) contributes to adaptively seeking out hard-to-distinguish predicates, which progressively sets learning targets and optimizes model's learning process with fine-grained supervision.
Furthermore, as predicate correlations vary with contexts of entities, the Adaptive Entity Discriminating Loss (EDL-A) adaptively adjusts the discriminating process under predictions of entities.
Utilizing CDL-A and EDL-A, our method can determine whether predicate pairs are hard-to-distinguish or not during training, which guarantees a more balanced and efficient learning process than previous methods \cite{cogtree},~\cite{prior:iccv},~\cite{pcpl},~\cite{bgnn}.
Finally, as the current evaluation metrics cannot comprehensively reflect model's capability of discriminating among hard-to-distinguish predicates, we introduce the Discriminatory Power (DP) to evaluate SGG models' discriminatory ability among hard-to-distinguish predicates.

This paper reinforces the preliminary version of our work~\cite{scenegraph:fgpl} with an effective optimization mechanism and insight analysis of each key component. 
The extensive contributions within this paper are fourfold.
Firstly, we enrich the related works with additional details on more literatures from three relevant research areas, i.e., scene graph generation, long-tailed distribution classification and fine-grained image classification.
Secondly, we propose an adaptive version of FGPL (i.e., FGPL-A) which can better differentiate among hard-to-distinguish predicates based on more comprehensive understanding of predicate correlations.
In practice, we design a Batch-Refinement (BR) regime which iteratively refines predicate correlations of the PL-A in keeping with model's dynamic learning status, providing faithful guidance for predicates discrimination.
Thirdly, by exploring the refined predicate correlations, the CDL-A progressively optimizes model's learning process with fine-grained supervision, while the EDL-A adaptively selects hard-to-distinguish predicates for sufficient regularization concerning model's current learning status. 
Taking advantage of both CDL-A and EDL-A, our proposed FGPL-A ensures a more balanced and efficient learning procedure compared to its original version, i.e., FGPL.
Finally, we present a more comprehensive experimental evaluation of FGPL and FGPL-A on two SGG datasets (i.e., VG-SGG and GQA-SGG) and validate the practicability of our methods on Sentence-to-Graph Retrieval and Image Captioning tasks.

Our main contributions are summarized as follows: 
\begin{itemize}
	\item Beyond the long-tailed problem in SGG tasks, we are the first to propose the fine-grained predicates problem, a new perspective of cases that hamper current SGG models. We further propose a novel plug-and-play Adaptive Fine-Grained Predicates Learning (FGPL-A) framework to well address this problem.
	\item We devise an Adaptive Predicate Lattice (PL-A) to adaptively explore pair-wise predicate correlations based on model's ongoing predictions of the current mini-batch, which provides faithful guidance for model's discriminating process. The Adaptive Category Discriminating Loss (CDL-A) helps SGG models learn to figure out and differentiate among hard-to-distinguish predicates with fine-grained supervision. Moreover, the Adaptive Entity Discriminating Loss (EDL-A) adaptively adjusts the discriminating process under predictions of entities, ensuring a balanced and efficient learning procedure. 
	\item Extensive experimental results demonstrate that our FGPL-A dramatically boosts the performance of benchmark models on the VG-SGG and GQA-SGG datasets (e.g., Transformer, VCTree, and Motif improved by \textbf{142.3$\%$, 175.2$\%$, 157.6$\%$ of Mean Recall (mR@100)} under the Predicate Classification sub-task on VG-SGG dataset), reaching a new state-of-the-art performance. Furthermore, we conduct experiments on Sentence-to-Graph Retrieval and Image Captioning tasks, which demonstrate the practicability of our FGPL-A.
\end{itemize}

\section{Related work}
\label{sec:related work}
\subsection{Scene Graph Generation}
Suffering from the biased prediction, today's SGG task is far from practical. To deal with the problem, some methods \cite{ssg:vctree,imp,motifs,scenegraph:cte} are proposed to refine the relation representation by exploring the contextual information via message passing.
Motif \cite{motifs} acquires the rich context representation among objects and relationships by encoding their features utilizing Bi-LSTMs.
For VCTree\cite{ssg:vctree}, it adaptively adjusts the structure of entities within the images to obtain the hierarchical information with a dynamic tree structure.
Taking advantage of self-attention mechanism, Transformer \cite{networks:transformer} explores the interactions among instances and their relationships, which contributes to refining the insufficient context information within each entity.  
Moreover, \cite{bgnn} proposes a bipartite graph neural network to capture the dependency between entities and predicates.
Other methods~\cite{tde,mr,bgnn,prior:iccv,pcpl} are proposed to balance the discriminating process in accordance with class distribution or visual clues.
\cite{bgnn} proposes a bi-level re-sampling scheme to achieve a balanced data distribution for training.
Besides, TDE~\cite{tde} attempts to disentangle predicates' unbiased representation from biased predictions. 
Concerning the consistency in scene graphs, \cite{energy} makes SGG models aware of the structural information in output space to constrain the predicates prediction. 
\cite{cogtree} explores predicate correlations with a hierarchical cognitive tree, in which model's learning process is regularized in a coarse-to-fine manner.
Additionally, \cite{pcpl} designs a global structure to explore predicate correlations for balancing model's learning process.
While the correlation among predicates varies with contexts, it is neither hierarchical nor global defined in \cite{pcpl,pcl,cogtree}. Different from them, we focus on discriminating among hard-to-distinguish predicates with pair-wise predicate correlations, constructed as a predicate graph.

\subsection{Long-Tailed Distribution Classification}
The long-tailed distribution reveals the nature of the real-world that a small number of classes dominates the majority of our observations. 
Recently, the long-tailed problems received growing attentions.
To solve the long-tailed problem, various distribution-based re-balancing learning strategies~\cite{adaptive,eqb,eql,group_softmax,logits,rethink,seesaw,tang_long_tail,balanced_loss} have been proposed. 
With respect to the imbalanced distribution, \cite{balanced_loss} has been proposed to re-balance the optimizing process according to the class-wise frequency. 
Considering the over-suppression problem,~\cite{eql,seesaw} reduces overwhelming punishment from head classes.
As a re-weighting method, Focal Loss \cite{focal} is proposed to adaptively adjust the loss weights to model's learning quality.
To keep a balanced learning process, \cite{seesaw} regularizes the learning process in reference to both model's prediction results and class distribution, which copes with the over-confidence problem on tail classes.
Similarly, \cite{adaptive} adaptively adjusts the training process for different classes on the basis of probabilities concerning the discriminatory difficulty between different classes. 
Furthermore, \cite{eqb} adapts the training procedure to model's learning status, which ensures a robust optimization process.
However, due to the considerable semantic overlap between different predicates, the predicate correlations are crucial for differentiating among hard-to-distinguish predicates in scene graph generation task. Therefore, in this work, we take advantage of both predicate distribution and predicate correlations to handle this issue.

\begin{figure*}[t]
\centering
\includegraphics[width=0.8\textwidth]{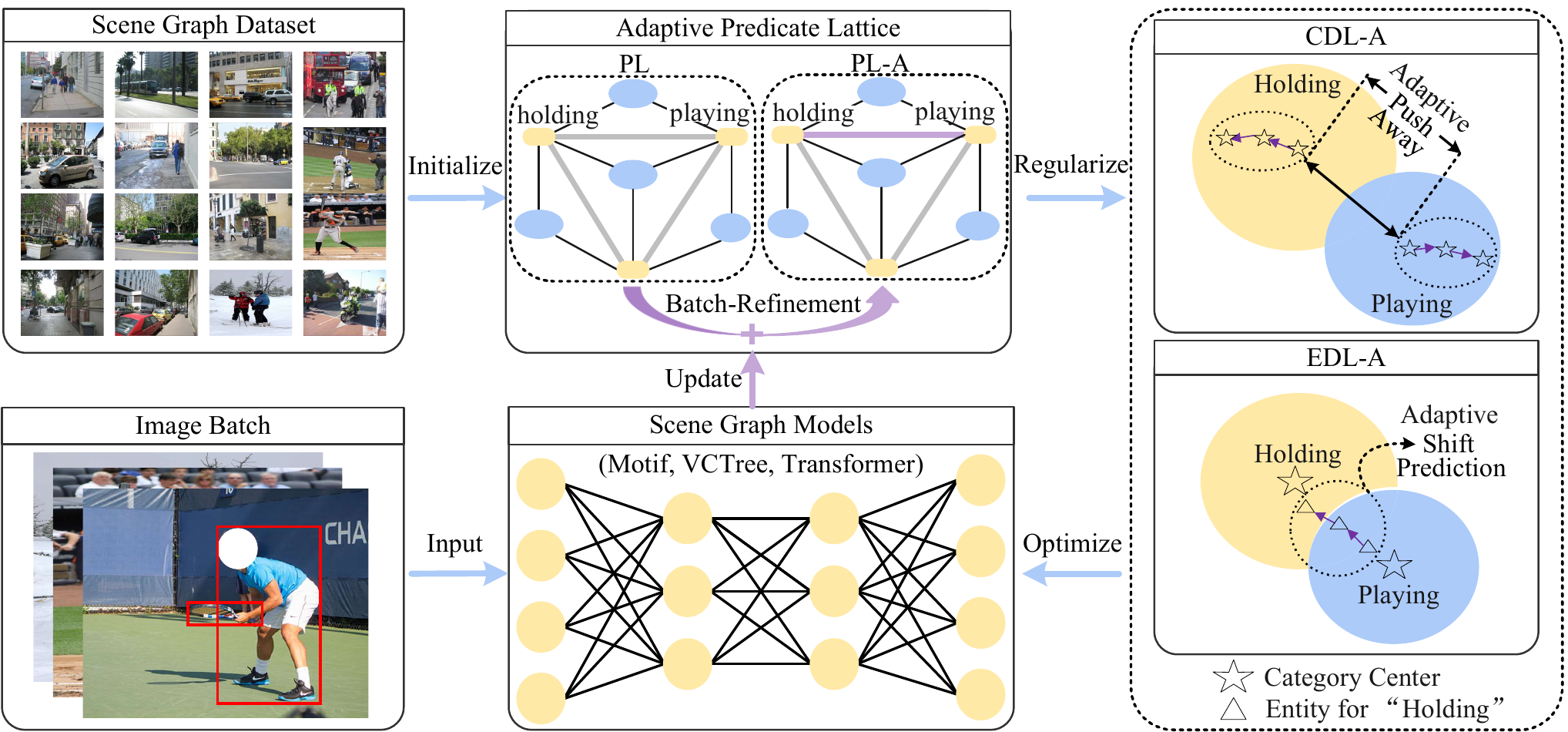}
\caption{\textbf{The Overview of the Adaptive Fine-Grained Predicates Learning (FGPL-A) framework.} It includes three parts: Adaptive Predicate Lattice (PL-A), Adaptive Category Discriminating Loss (CDL-A), and Adaptive Entity Discriminating Loss (EDL-A). FGPL-A can be incorporated into several benchmark SGG models. The Adaptive Predicate Lattice (PL-A) is initialized as the PL, and then gets iteratively refines by the Batch-Refinement (BR) based on model's predictions of the current mini-batch. By exploring the refined predicate correlations of PL-A, CDL-A and EDL-A adaptively optimize SGG models, ensuring a balanced and efficient leaning process.}
\label{fig:framework_FGPL_A}
\end{figure*}

\subsection{Fine-Grained Image Classification}
Fine-Grained Image Classification aims to recognize hard-to-distinguish objects in a coarse-to-fine manner. Existing methods tackle the problem roughly from two perspectives, representation-encoding~\cite{fg5,fg7,fg8} and local recognition~\cite{fg1,fg3,fg4}. The former~\cite{fg5,fg7,fg8} explore visual clues for fine-grained classification, while the latter~\cite{fg1,fg2,fg3,fg4} figure out recognizable parts for discriminating. 
However, due to complex relationships among predicates, such a coarse-to-fine discriminatory manner may fail to differentiate predicates for scene graph generation. Particularly, different predicates may share similar meanings in a specific scenario, while a predicate may have different meanings in different contexts. As a result, instead of a hierarchical structure, predicate correlations should be formed as a graph structure. Concretely, we construct a Predicate Lattice to comprehend predicate correlations for predicate discriminating.

\section{Adaptive Fine-Grained Predicates Learning}
\label{sec:FGPL}
\subsection{Problem Formulation and Overview}
\noindent\textbf{Problem Formulation}:
Scene graph generation is typically a two-stage multi-class classification task. In the first stage, Faster R-CNN~\cite{obj_det:faster,obj_det:rcnn} detects instance labels $O=\{o_i\}$, bounding boxes $B=\{b_i\}$, and feature maps $X=\{x_i\}$ within an input image $I$. In the second stage, scene graph models infer predicates from subject $i$ to object $j$, i.e., $R=\{r_{i,j}\}$, based on detection results, i.e., $Pr(R|O,B,X)$.

\noindent\textbf{Overview of FGPL framework:}
Within our Fine-Grained Predicates Learning framework (FGPL)~\cite{scenegraph:fgpl}, we first construct a Predicate Lattice (PL) concerning context information to understand ubiquitous correlations among predicates. Then, utilizing the PL, we develop a Category Discriminating Loss (CDL) and an Entity Discriminating Loss (EDL) which help SGG models differentiate among hard-to-distinguish predicates.

\noindent\textbf{Limitations of FGPL:}
Obtained from models' biased predictions of SGG dataset before training, the predicate correlations of the PL are pre-determined and static during training.
Regularized by the PL-based CDL and EDL, FGPL fails to cope with model's dynamic learning pace, rendering over-confidence~\cite{prob:overconfident,overconfident} on predicate discrimination.

\noindent\textbf{Overview of FGPL-A framework:}
Based on the above observations, we further devise an adaptive version for FGPL as Adaptive Fine-Grained Predicate Learning (FGPL-A) by introducing an Adaptive Predicate Lattice (PL-A), which progressively updates predicate correlations in keeping with model’s dynamic learning status.
Practically, we propose a Batch-Refinement (BR) regime for the PL-A, which iteratively refines pair-wise predicate correlations via exploring model's ongoing predictions of the current mini-batch in training iterations. 
Then, by exploring the refined predicate correlations, we propose the adaptive version of CDL and EDL as Adaptive Entity Discriminating Loss (EDL-A) and Adaptive Entity Discriminating Loss (CDL-A) for the adaptive discrimination among hard-to-distinguish predicates, resulting in a balanced and efficient learning process. 
The overview of FGPL-A framework is shown in Fig.~\ref{fig:framework_FGPL_A}.


\subsection{Adaptive Predicate Lattice Construction}
In this section, we describe the construction process of Adaptive Predicate Lattice (PL-A), which provides comprehensive understanding on predicate correlations, concerning contexts and model's dynamic learning status.
First, we build the Predicate Lattice (PL) by exploring model's biased predictions of SGG dataset in Sec.~\ref{sec:construction}.
Since the PL fails to cope with model's dynamic learning status, we further devise an Adaptive Predicate Lattice (PL-A)  by introducing a Batch-Refinement (BR) regime in Sec.~\ref{sec:PLA}.

\subsubsection{Predicate Lattice}
\label{sec:construction}
\begin{figure}[t]
\centering
\includegraphics[width=0.45\textwidth]{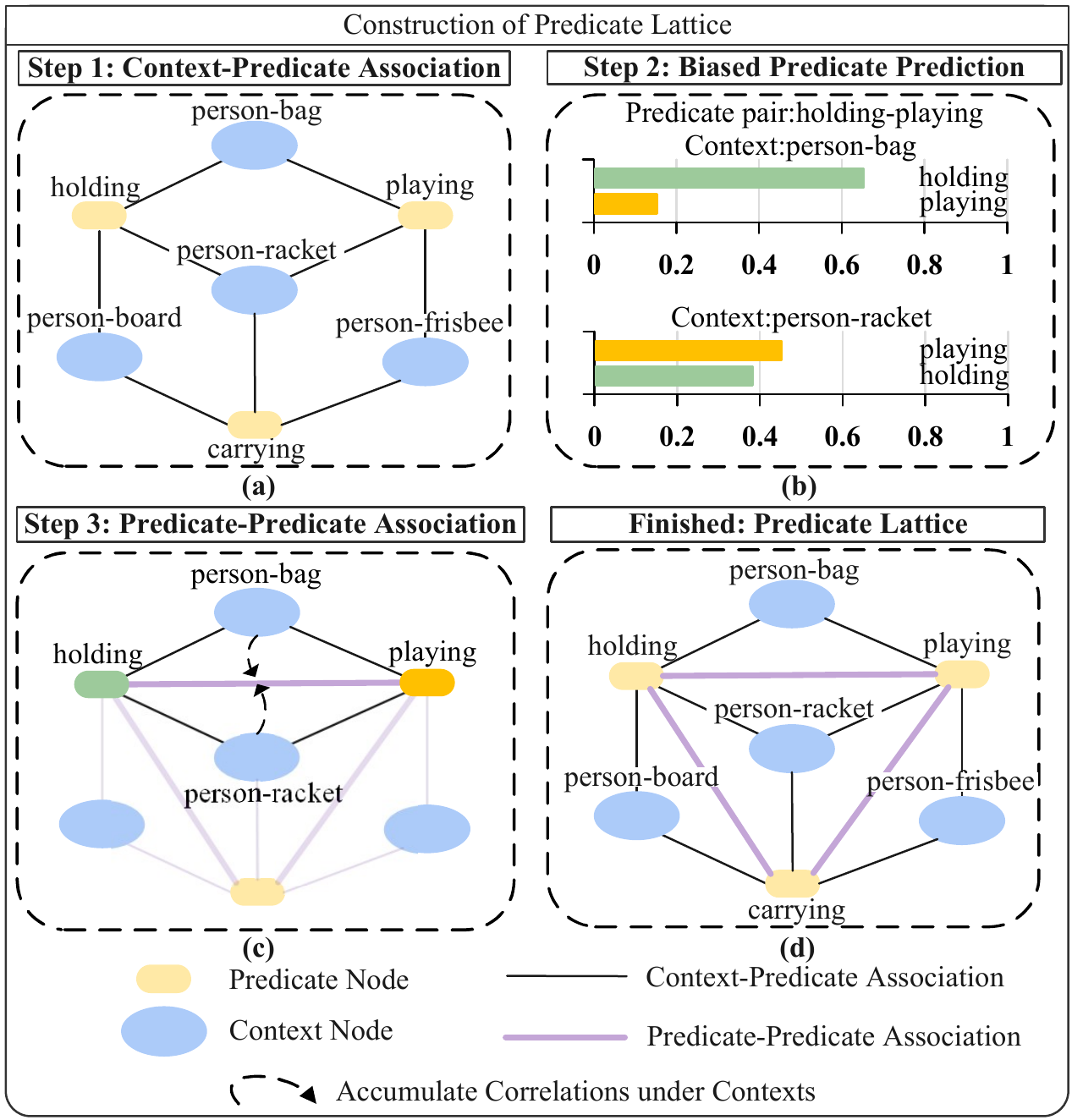} 
\caption{\textbf{Construction of Predicate Lattice.} The whole process is divided into three steps: (1) Context-Predicate Association; (2) Biased Predicate Prediction; and (3) Predicate-Predicate Association.}
\label{fig:lattice}
\end{figure}

To fully understand relationships among predicates, we build a Predicate Lattice (PL), which includes correlations for each pair of predicates concerning contextual information. In general, predicate correlations are acquired under different contexts, since contexts (i.e., visual or semantic information of predicates’ subjects and objects) determine relationships among predicates. Specifically, we extract their contextual-based correlations from biased predictions containing all possible contexts between each pair of predicates. As shown in Fig.~\ref{fig:lattice}, the construction procedure can be divided into three steps:

\noindent\textbf{Context-Predicate Association}: 
We first establish Context-Predicate associations between predicate nodes and context nodes. As contexts determine correlations among predicates, predicate correlations are constructed as a Predicate Lattice containing predicates and related contexts (i.e., visual or semantic information of predicates' subjects and objects). In Fig.~\ref{fig:lattice}(a), we show structures of our Predicate Lattice. There are two kinds of nodes in Predicate Lattice, namely Predicate nodes and Context nodes, which indicate predicate categories and labels of subject-object pairs, respectively. Several predicate nodes connect to the same context node, which denotes that several predicates can describe relationships in the same context. For instance in Fig.~\ref{fig:lattice}(a), both ``holding'' and ``carrying'' can be utilized to describe relationships for ``person-racket''. Specifically, we adopt Frequency model~\cite{motifs} to derive every subject-object pair as the context for each predicate from the SGG dataset (Visual Genome). Moreover, weights of edges between predicate nodes and context nodes, i.e., $Pr(r_{i,j}|o_i,o_j)$, denote the occurrence frequency for each ``subject($o_i$)-predicate($r_{i,j}$)-object($o_j$)'' triplet in dataset. In this way, we establish connections between predicate nodes and context nodes in Predicate Lattice.

\noindent\textbf{Biased Predicate Prediction}:
To associate predicate pairs with predicate correlations in the next step, we acquire Biased Predicate Prediction from SGG models. Firstly, we incorporate Context-Predicate Association, constructed in step one, into SGG models. 
Particularly, we extract the Context-Predicate Association for each ``subject-predicate-object'' triplet as semantic information. Then, to acquire complete contextual information, we combine semantic information with visual features, i.e., $B={b_i}$ and $X$, of subjects $o_i$ and objects $o_j$ to predict predicates $Pr(r_{i,j}|o_i,o_j,b_i,b_j,x_i,x_j)$. With the contextual information, we derive the Biased Predicate Prediction of pre-trained SGG models by inferring on the training set of the SGG dataset concerning all scenarios. In this way, the Biased Predicate Prediction contains predicate predictions under all possible scenarios for each predicate pair. For instance, as shown in Fig.~\ref{fig:lattice}(b), we infer the pre-trained SGG model under all possible scenarios for predicate ``playing'' or ``holding'', such as ``person-racket'' and ``person-bag''.

\noindent\textbf{Predicate-Predicate Association}: At last, we establish Predicate-Predicate Association among predicates with context-based correlations obtained from the Biased Predicate Prediction. 
The Biased Predicate Prediction implies the context-based correlations between each pair of predicates.
For instance, if most samples are predicted as $j$ but labeled as $i$ in ground truth, predicate $i$ is correlated to predicate $j$ in most contexts.
Based on the above observation, we accumulate prediction results from each possible context to obtain the holistic predicate correlations between each pair of predicates, shown in Fig.~\ref{fig:lattice}(c). 
For instance, given predicate pair ``playing-holding'', we gather their correlations under all contexts/scenarios, such as ``person-racket'' and ``person-bag''. Moreover, if predicate $i$ is correlated to predicate $j$ in most contexts, they are prone to be strongly correlated. Therefore, we normalize the gathered predicate correlations as $S=\{s_{i,j}\}$ with $s_{i,j}$, which indicates the proportion of samples labeled as $i$ but predicted as $j$. In particular, higher $s_{i,j}$ means a more substantial correlation between predicate pair $i$ and $j$. Then, we associate predicate pairs with predicate correlations $s_{i,j}$. Finally, predicate correlations are formed as a Predicate Lattice, shown in Fig.~\ref{fig:lattice}(d).

\subsubsection{Adaptive Predicate Lattice with Batch-Refinement}
\label{sec:PLA}
\noindent\textbf{Limitations of PL:} Ignoring model's dynamic learning status, the PL, built in Sec.~\ref{sec:construction}, fails to accurately monitor model's learning quality among predicates (i.e., hard-to-distinguish or recognizable) throughout training.
Derived from the biased predictions before training, the predicate correlations of PL are pre-determined and remain static thereafter. 
As the discriminatory power are progressively acquired, the static predicate correlations of PL are unable to catch the variations of model's learning status during the learning procedure.
Furthermore, regularized by the intrinsic predicate correlations, model's learning process tends to be inconsistent with its gradually gained discriminatory power, rendering the over-confident issue on predicate discrimination.
Thus, explorations on the refinement mechanism of predicate correlations need to be further carried out for a balanced and efficient learning process.

\noindent\textbf{Formulation of PL-A:}
To address the limitation mentioned above, we further extend the Predicate Lattice (PL) as Adaptive Predicate Lattice (PL-A).
Practically, to make predicate correlations in line with model's gradually obtained discriminatory power, we devise a Batch-Refinement (BR) regime, which iteratively refines pair-wise predicate correlations of PL-A via exploring model's ongoing predictions.
For Batch-Refinement (BR) regime, it first comes to mind to iteratively refine predicate correlations by acquiring model's biased predictions of the whole training set at each iteration.
However, obtaining evaluation results for the whole training set at each training step is infeasible, which brings a tremendous computational cost. 
Instead, we iteratively update the predicate correlations with the accumulated predictive scores for each mini-batch, since the batch-based predictions approximately reflect model's learning status at each learning pace.

\begin{figure}[t]
\begin{center}
\includegraphics[width=0.46\textwidth]{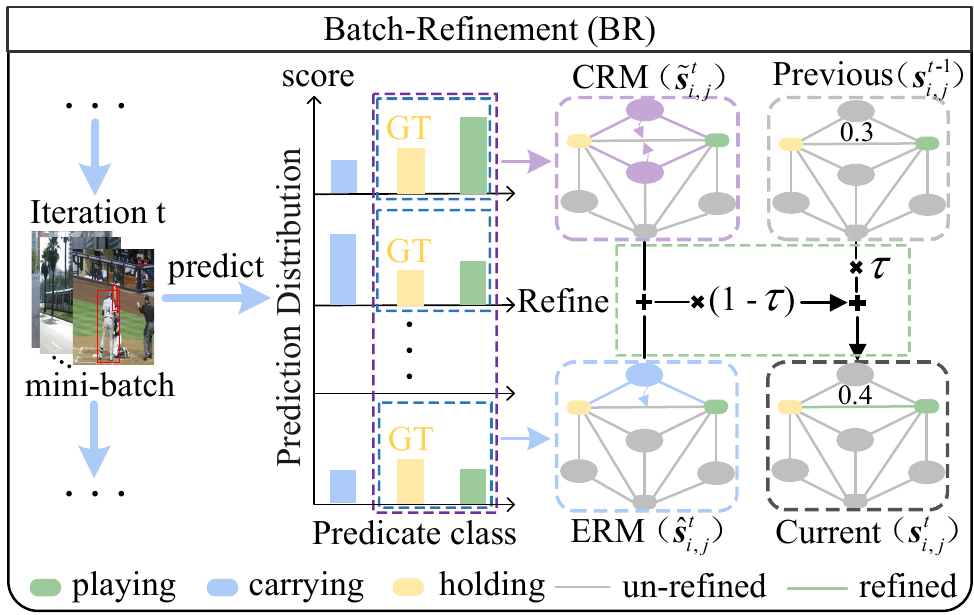}
\caption{\textbf{The illustration of Batch-Refinement (BR) regime.} BR refines the predicate correlations with Category Refining Momentum (CRM) and Entity Refining Momentum (ERM). The Category Refining Momentum (CRM) updates the predicate correlations by jointly accumulating the mis-classification results within each mini-batch $M(m=0,1,2...)$ at category-level. The Entity Refinement Momentum (ERM) updates the predicate correlations by individually exploring the biased predictions concerning the specific context within each sample at entity-level.}
\label{fig:BR}
\label{BR}
\end{center}
\end{figure}

In practice, the PL-A is initialized with the predicate correlations (${\rm{s}}_{i,j}$) of the PL at first, since PL provides fundamental insights on comprehending the predicate correlations.
Then, for the $t$-th iteration, we update the pair-wise predicate correlations of PL-A by accumulating models' biased predictions within the current mini-batch utilizing the Batch-Refinement (BR) regime. 
The corresponding Batch-Refinement (BR) regime can be written as follows:

\begin{equation}
	\begin{aligned}
	&{\rm{s}}_{i,j}^t=\begin{cases}
    \tau~{\rm{s}}_{i,j}^{t - 1} + (1-\tau)~{\rm{S}}_{i,j}^t ,  & \text{ if } t > 0 \;  \\
    {\rm{s}}_{i,j}, & \text{ if } t = 0 \; \\
    \end{cases}  ~, \\
    &{\rm{S}}_{i,j}^t =  {\rm{\tilde s}}_{i,j}^t + {\rm{\hat s}}_{i,j}^t, \\
	\end{aligned}
			\label{Eq:BR}
\end{equation}

where ${\rm{s}}_{i,j}^t$ and ${\rm{s}}_{i,j}^{t - 1}$ indicates the predicate correlations between class $i$ and $j$ at the $t$-th and the $t$-1 th iteration. 
Moreover, ${\rm{s}}_{i,j}^t$ is refined by iteratively adding the Refining Momentum ${\rm{S}}_{i,j}^t$ to ${\rm{s}}_{i,j}^{t-1}$. 
Intuitively, ${\rm{s}}_{i,j}^{t - 1}$ and ${\rm{S}}_{i,j}^t$ reveals the historical and and coming-batch statistics of predicate correlations.
Besides, $\tau$ denotes the hyper-parameter for a trade-off between ${\rm{s}}_{i,j}^{t - 1}$ and  ${\rm{S}}_{i,j}^t$.  
Moreover, the Refining Momentum ${\rm{S}}_{i,j}^t$ is composed of both Entity Refining Momentum (ERM) ${\rm{\hat s}}_{i,j}^t$ and Category Refining Momentum (CRM) ${\rm{\tilde s}}_{i,j}^t$.
Practically, CRM updates the predicate correlations by jointly accumulating the mis-classification results between the positive and negative classes within each mini-batch $M(m=0,1,2...)$.
As the CRM only takes the inter-class correlation into account, the remarkable intra-class variance of predicates, varying with contexts, gets diminished in the normalization operation of the category-level refinement.
Hence, as the complement for CRM, we further design an ERM which refines the predicate correlations by individually exploring the mis-classification concerning the specific context within each sample.
The illustration of the CRM and ERM is shown in Fig.~\ref{fig:BR}.

Practically, the Entity Refining Momentum (ERM) is computed as:
\begin{equation}
	\begin{aligned}
	& {\rm{\hat s}}_{i,j}^t = \frac{{p_j^{t,m}}}{{p_i^{t,m}}},
	\end{aligned}
\end{equation}
where the ERM ${\rm{\hat s}}_{i,j}^t$ investigates the predicate correlations within different samples concerning the intra-class diversity of contextual information.
In specific, it individually calculates the ratio between predictive scores of negative class $j$ and positive class $i$ within in sample $m$ at iteration $t$, i.e., ${p_j^{t,m}}$ and ${p_i^{t,m}}$.
Intuitively, $\frac{{p_j^{t,m}}}{{p_i^{t,m}}}$ reveals how likely the given sample $m$ would be misclassified as a negative class $j$ at $t$ -th iteration by SGG models under the specific context.

Additionally, the Category Refining Momentum (CRM) is formulated as follows:
\begin{equation}
	\begin{aligned}
    & {\rm{\tilde s}}_{i,j}^t = \frac{{\sum\nolimits_{m = 1}^M {p_j^{t,m}} }}{{\sum\nolimits_{m = 1}^M {p_i^{t,m}} }},\\
	\end{aligned}
\end{equation}
where CRM ${\rm{\tilde s}}_{i,j}^t$ fully acquires the inter-class predicate correlations by jointly exploring the biased predictions among different categories, accumulated from a batch of samples $M$ at $t$-th iteration.
Concretely, it first gathers the holistic prediction distribution of negative class $j$ and positive class $i$ from $M$ samples in the $t$-th mini-batch, as ${\sum\nolimits_{m = 1}^M {p_j^{t,m}}}$ and ${\sum\nolimits_{m = 1}^M {p_i^{t,m}}}$. 
Then, the CRM ${\rm{\tilde s}}_{i,j}^t$ is derived by normalizing scores of negative class $j$ conditioned on that of the positive one $i$ , i.e., $ \frac{{\sum\nolimits_{m = 1}^M {p_j^{t,m}} }}{{\sum\nolimits_{m = 1}^M {p_i^{t,m}} }}$, which reveals the difficulty of distinction between the positive category $i$ and the negative class $j$ for SGG models at current stage. Refined in terms of the CRM (inter-class) and the ERM (intra-class), the predicate correlations in the PL-A provide a faithful guidance (i.e., predicate pairs are hard-to-distinguish or recognizable) for the Category Discriminating Loss (CDL) and the Entity Discriminating Loss (EDL), which are elaborated in the following sections.



\subsection{Adaptive Category Discriminating Loss}
\label{sec:CDL}
To compensate for the limitations of the re-weighting method, we introduce our Category Discriminating Loss (CDL) in Sec.~\ref{sec:CDL}, which attempts to differentiate among hard-to-distinguish predicates while maintaining the discriminatory power on distinguishable ones.
Since the CDL fails to cope with model's dynamic learning status, we further devise an Adaptive Category Discriminating Loss (CDL-A) in Sec.~\ref{sec:CDL-A}, which contributes to achieving more substantial discriminatory power among predicates concerning model's dynamic learning pace.

\subsubsection{Category Discriminating Loss}
\label{sec:CDL}
\noindent\textbf{Limitations of Re-weighting Method:}
Overall, recent re-weighting methods re-balance the learning process by strengthening the penalty to head classes while scaling down the overwhelming punishment to tail classes. Specifically, the state-of-the-art re-weighting method~\cite{seesaw} adjusts weights for each class in Cross-Entropy Loss on the basis of the proportion of training samples as follows:

\begin{equation}
\label{balanced_softmax}
	\begin{aligned}
    &\mathcal{L}_{CE}(\eta) = -{\textstyle \sum_{i=1}^{C}}y_ilog(\hat{\phi}_i)~,\\
    &\hat{\phi}_i = \frac{e^{\eta_i}}{ {\textstyle \sum_{j=1}^{C}w_{i,j}e^{\eta_j}}}~,
    w_{i,j}=\begin{cases}
     {(\frac{{{n_j}}}{{{n_i}}})^\alpha },  & \text{ if } n_j > n_i \;  \\
    1, & \text{ if } n_j \le n_i \; \\
    \end{cases}  ~, \\
	\end{aligned}
\end{equation}
where $\eta=[\eta_1,\eta_2,...,\eta_C]$ and $\hat\phi=[\hat\phi_1,\hat\phi_2,...,\hat\phi_C]$ denote predicted logits and re-weighted probabilities for each class. The label $Y=[y_1,y_2,...,y_C]$ is a one-hot vector. Additionally, $w_{i,j}$ denotes the re-weighting factor concerning distribution between positive class $i$ and negative class $j$. Explicitly, $w_{i,j}$ is calculated based on the proportion of distribution between class $i$ and $j$, as shown in Eq.~\ref{balanced_softmax}, where $\alpha > 0$. 
\begin{equation}
	\begin{aligned}
\frac{\partial \mathcal{L}_{CE}(\eta)}{\partial \eta_j} = \frac{w_{i,j}e^{\eta_j }}{ {\textstyle \sum_{k=1}^{C}w_{i,k}e^{\eta_k}}}~.
	\end{aligned}
\label{eq.grad}
\end{equation}
Eq~.\ref{eq.grad} shows negative gradients for category $j$. If positive category $i$ is less frequent than negative category $j$, i.e., $n_j > n_i$ with $w_{i,j}>1$, it will strengthen the punishment to negative class $j$. On the contrary, if $n_j \le n_i$ with $w_{i,j}=1$, it will degrade the penalty to negative class $j$. Finally, it results in a balanced learning process.

Without considering predicate correlations, re-weighting methods cannot adaptively adjust discriminating process in accordance with difficulty of discrimination, resulting in an inefficient learning process. 
As an inherent characteristic of predicates, predicate correlations reveal difficulty of discrimination for different pairs of predicates. 
However, ignoring predicate correlations in learning process, the re-weighting method roughly reduces negative gradients for all negative predicates with fewer samples than the positive predicate.
As a process to push away the decision boundary from tail classes to head classes, such discriminating process is prone to over-suppress weakly correlated predicate pairs and degrades the learned discriminatory capability of recognizable predicates as maintained in \cite{adaptive,eqb}. 
Take an example among ``on/has/\underline{standing on}'', where ``on-\underline{standing on}'' are strongly correlated and ``has-\underline{standing on}'' are weakly correlated. To prevent the tail class ``standing on'' from being over-suppressed, the re-weighting method roughly degrades negative gradients from both ``on'' and ``has''. Although it strengthens discriminatory power between ``on'' and ``standing on'', it is prone to reduce that between ``has'' and ``standing on'' simultaneously. 

\noindent\textbf{Formulation of CDL:}
Based on the above observations, we should both consider the class distribution and predicate correlations to differentiate among hard-to-distinguish predicates. Thus, based on the re-weighting method in Eq.~\ref{balanced_softmax}, we devise Category Discriminating Loss (CDL), which adjusts the re-weighting process according to predicate correlations obtained from Predicate Lattice.
Overall, we utilize predicate correlations $s_{i,j}$, defined in Sec.~\ref{sec:construction}, as a signal to adjust the degree of re-weighting between predicates $i$ and $j$.
Especially, we mitigate the magnitude of re-weighting for weakly correlated predicates while strengthening that for strongly correlated ones by setting $w_{i,j}$, in Eq.~\ref{balanced_softmax}, with different values.
In this way, we maintain gained discriminatory power among recognizable predicates and further enhance that among hard-to-distinguish ones, shown as below:

\begin{equation}
	\begin{aligned}
w_{i,j}&=\begin{cases}
 \mu_{i,j}^{\beta}~(\ge 1),  & \text{ if } \mu_{i,j}\ge 1 \; and  \; \varphi_{i,j}  >\xi~  \\
 1, & \text{ if } \mu_{i,j}\ge 1 \; and  \; \varphi_{i,j}  \le \xi~ \\
 1, & \text{ if } \mu_{i,j} < 1 \; and  \; \varphi_{i,j}  >\xi~   \\ 
 \mu_{i,j}^\alpha~(< 1),  & \text{ if }\mu_{i,j} < 1 \; and  \; \varphi_{i,j} \le \xi~  \\
\end{cases}  ~, \\
\mu_{i,j} &= \frac{n_{j}}{n_{i}}, ~\varphi_{i,j} = \frac{s_{i,j}}{s_{i,i}},
	\end{aligned}
		\label{predicate_correaltion}
\end{equation}
where $\varphi_{i,j}$ is calculated by the proportion between $s_{i,j}$ and $s_{i,i}$, revealing correlations between predicate $i$ and $j$. In addition, $\alpha$ and $\beta$ are hyper-parameters larger than $0$.
For instance, when $n_j \ge n_i$ ($\mu_{i,j} \ge 1$), if $\varphi_{i,j}>\xi$ of strongly correlated predicate pair $i$ and $j$, $w_{i,j}$ is larger than $1$ to strengthen the punishment on negative predicate $j$. 
In contrast, if $\varphi_{i,j}\le\xi$ of weakly correlated predicate pair $i$ and $j$, $w_{i,j}$ is set as $1$ to mitigate the magnitude of penalty on negative predicate $j$. That is because the excessive punishment is unnecessary for the weakly correlated predicate $j$, which is easy to distinguish from predicate $i$ for models.
When $n_j < n_i$ ($\mu_{i,j} < 1$), we set $w_{i,j}\le1$ (including $\varphi_{i,j} > \xi$ and $\varphi_{i,j} \le \xi$) to relieve the over-suppression from head predicate $i$ to tail one $j$. Moreover, if $\varphi_{i,j} \le \xi$, we set $w_{i,j} = \mu_{i,j}^\alpha~(< 1)$ to mitigate the magnitude of the penalty on negative predicate $j$.

\subsubsection{Adaptive Category Discriminating Loss}
\label{sec:CDL-A}
\noindent\textbf{Limitations of CDL:}
Founded on the Cross-Entropy Loss, the CDL proposed in Sec.~\ref{sec:CDL} has two limitations.
Firstly, the hard targets (one-hot encoded) in CDL roughly regularize model's learning process in a coarse-grained manner.
The coarse-grained supervision is inadequate for the fine-grained predicates learning, which inevitably introduces huge ambiguity into model's learning procedure.
Thus, fine-grained supervision ought to be applied for regularizing the fine-grained discriminating process.
Secondly, ignoring model's dynamic learning status, the predetermined and static label distribution has been found to incur an overconfident problem~\cite{prob:overconfident,overconfident}. Consequently, it may lead to degradation in model's discriminatory power and dramatically limits its learning efficacy.
Hence, we claim that different label distributions should be progressively assigned for each category in keeping with model's learning pace over the training course.

\noindent\textbf{Formulation of CDL-A:}
To compensate for the limitations mentioned above, we adaptively set learning targets utilizing the fine-grained predicate correlations for SGG models.
To achieve that, we introduce an Adaptive Label Softening (ALS) scheme.
The ALS dynamically softens hard targets with refined predicate correlations of the PL-A.
As the refined predicate correlations reveal model's learning quality among different categories at current stage, the progressively softened targets guarantee a balanced and efficient learning process for SGG models.
Moreover, the softened targets provide insightful understanding of predicate correlations during training, which regularize the learning process with fine-grained supervision, enhancing models' discriminatory power among hard-to-distinguish predicates. The corresponding Adaptive Category Discriminating Loss (CDL-A) can be written as follows:
\begin{equation}
	\begin{aligned}
&{\mathcal{L}_{CDA}}(\eta ) =  \mathcal{L}_{CD}(\eta )+ \theta{\mathcal{L}_{ALS}}(\eta )~,\\
	& {\mathcal{L}_{ALS}}(\eta )= -\sum\nolimits_{i = 1}^C { s_{i,j}^t} \log ({\hat \phi _i}) ~,\\
	\end{aligned}
				\label{eq:CDL_A}
\end{equation}
where $\mathcal{L}_{ALS}$ and $\mathcal{L}_{CD}$ denotes the Adaptive Label Softening (ALS) regularization and the Category Discriminating Loss (CDL). It is worth noting that, since the CDL provides fundamental effectiveness in differentiating among hard-to-distinguish predicates, we treat the ALS (${\mathcal{L}_{ALS}}$) as the regularization for CDL, with $\theta$ as the trade-off coefficient. Additionally, ${\hat s_{i,j}^t}$ and ${\hat \phi _i}$ denotes the refined predicate correlations at the $t$-th iteration mentioned in Sec.~\ref{sec:PLA}, and the re-weighted logits for class $i$ in Sec.~\ref{sec:CDL}. 
In contrast to hard targets (one-hot encoded) in CDL, CDL-A progressively optimizes model's learning process with fine-grained supervision (softened learning objectives, i.e., ${ s_{i,j}^t}$).
Under the guidance of fine-grained objectives, CDL-A enhances models' capability for fine-grained discrimination among hard-to-distinguish predicates.
Moreover, as predicate correlations ${\hat s_{i,j}^t}$ are iteratively refined in keeping with model's gradually gained discriminatory power, CDL-A adaptively adjusts learning targets for SGG models at each learning pace.
Hence, CDL-A fits model's learning status and results in a balanced and effective discriminating process.

\subsection{Adaptive Entity Discriminating Loss}
In this section, we introduce our Entity Discriminating Loss (EDL) in Sec.~\ref{sec:EDL}, which adapts the discriminating process to different contexts within entities.
Furthermore, to compensate for the constraints of EDL, we propose the Adaptive Entity Discriminating Loss (EDL-A) in Sec.~\ref{sec:EDL-A}, which adaptively adjusts the discriminating process to model's ongoing predictions of the current mini-batch.


\subsubsection{Entity Discriminating Loss}
\label{sec:EDL}
\noindent\textbf{Limitations of CDL:}
Although CDL effectively differentiates hard-to-distinguish predicates, it still has a limitation: it only considers inter-class difficulty of discrimination between predicates, but ignores intra-class difficulty varied with different contexts within entities.
As model's predictive scores reflects discrimination difficulty within the specific context, we individually treat prediction results of each sample as signals to adjust the decision boundary. 

\noindent\textbf{Formulation of EDL:}
Based on the observations, we propose an Entity Discriminating Loss (EDL), which adapts the discriminating process to the contexts within entities, shown as below:
\begin{equation}
	{\mathcal{L}_{ED}(\eta)} = {\frac{1}{{\left| {\mathcal{V}_i  } \right|}}\sum\limits_{j \in \mathcal{V}_i} {\max (0,{\phi_j} - {\phi_i} + \delta )\frac{{{n_j}}}{{{n_i}}}}~, } 
			\label{eq:EDL}
\end{equation}
where $\mathcal{V}_i$ is defined as a set of strongly correlated predicates selected in reference to predicate correlations $s_{i,j}$ in Predicate Lattice (PL). 
For each predicate category $i$, $M$ predicates with the highest $s_{i,j}$ in the Predicate Lattice are chosen to construct $\mathcal{V}_i$. 
Given the input sample $\eta$, $\phi_i$ and $\phi_j$ are the predicted probabilities for predicates $i$ and $j$.
Intuitively, $\phi_j-\phi_i$ implies the discrimination difficulty between predicates $i$ and $j$ of the specific context within sample $\eta$. 
The $\delta$ is a hyper-parameter, which denotes prediction margins for predicates. Furthermore, EDL is reduced to zero if predicate pairs are recognizable enough i.e., ${\phi_i} - {\phi_j} \ge \delta$. Moreover, we also adopt the balancing factor $\frac{n_j}{n_i}$ to alleviate imbalanced gradients between classes with fewer or more observations.

\subsubsection{Adaptive Entity Discriminating Loss}
\label{sec:EDL-A}
\noindent\textbf{Limitations of EDL:}
While the EDL proposed in Sec.~\ref{sec:EDL} takes the intra-class variance of contexts into account, its effectiveness is still limited by the pre-determined predicate correlations within the Predicate Lattice (PL).
Based on static predicate correlations of the PL, EDL only concentrates on differentiating a small set of strongly correlated predicates, which ignores the variations of predicate correlations during model's leaning process.
Hence, such episode-fixed regularization is prone to be inconsistent with model's gradually obtained discriminatory power, resulting in the degradation on model's performance.

\noindent\textbf{Formulation of EDL-A:}
To remedy such phenomenon, we further devise an adaptive version of EDL, namely the Adaptive Entity Discriminating Loss (EDL-A), which adaptively picks strongly correlated predicates in accordance with model's gradually obtained discriminatory power, concerning inherent contexts within entities.
The EDL-A is written as follows:
\begin{equation}
	\begin{aligned}
	& {\mathcal{L}_{EDA}(\eta)} = {\frac{1}{{\left| {\mathcal{\hat V}_i  } \right|}}\sum\limits_{j \in \mathcal{\hat V}_i} {\max (0,{\hat \phi_j} - {\hat \phi_i} + \delta )\frac{{{n_j}}}{{{n_i}}}}~, } \\
	& \mathcal{\hat V}_i = topK({\rm{s}}_{i,j|j \neq i}^t) ~,\\
	\end{aligned}
			\label{eq:EDL_A}
\end{equation}
where ${s_{i,j}^t}$ and ${\hat \phi _i}$ denote the refined predicate correlations at current stage $t$ mentioned in Sec.~\ref{sec:PLA} and the re-weighted logits for class $i$ in Sec.~\ref{sec:CDL}. 
Moreover, $\mathcal{\hat V}_i$ indicates a set of hard-to-distinguish (strongly correlated) predicates, which are dynamically chosen in reference to the refined predicate correlations ${s_{i,j}^t}$ of the PL-A.
In practice, $\mathcal{\hat V}_i$ is built by figuring out the predicates $j$ with the top-k strongest correlations ${s_{i,j}^t}$ to predicate $i$ at current stage $t$. 
As EDL-A adaptively chooses which category to suppress for each sample in training iterations, it adapts the discriminating process to both model's learning status and contexts of entities, resulting in a robust and efficient learning process.

\subsection{Adaptive Discriminating Loss For FGPL-A}
Including both CDL-A in Eq.~\ref{eq:CDL_A} and EDL-A in Eq.~\ref{eq:EDL_A}, the Adaptive Discriminating Loss for FGPL-A can be expressed as below:
\begin{equation}
	\begin{aligned}
	\mathcal{L}_{DA}(\eta)= \mathcal{L}_{CDA}(\eta)+\gamma \mathcal{L}_{EDA}(\eta)~,
	\end{aligned}
	\label{ADL}
\end{equation}
where $\gamma$ balances the regularization between CDL-A and EDL-A. By exploring the progressively refined predicate correlation, the Adaptive Category/Entity Discriminating Loss (CDL-A/EDL-A) adaptively collaborates category-level and entity-level optimization, which guarantees an efficient learning procedure and enhances models discriminatory power among hard-to-distinguish predicates.

\section{Experiments}
\label{sec:experiments}

\subsection{Experiment Setting}
\noindent\textbf{Dataset}: Following previous works~\cite{motifs,mr}, we evaluate our methods on the Visual Genome (VG-SGG) dataset. It contains $108k$ images with $75k$ and $37k$ objects and predicates classes. Due to the severely skewed class distributions within the VG-SGG, following~\cite{tde,motifs}, we adopt the widely used split for SGG. Under the setting, the VG-SGG dataset has $150$ object categories and $50$ relationship categories. Then, we further divide it into $70\%$ training set, $30\%$ testing set, and $5k$ images, selected from the training set, for validation. To further validate the generalizability of our methods on other SGG datasets, we also conduct experiments on a more challenging SGG dataset, namely GQA-SGG. In contrast to VG-SGG, it has more object categories (1704 classes) and predicate categories (311 classes), which contains more complex scenario information. Similar to the VG-SGG, we adopt 70-30 split for train ($75k$) and test set ($10k$), and further sample a small set ($5k$) for validation. For Sentence-to-Graph Retrieval task, we follow \cite{tde} to sample the overlapped $41k$ images between VG and MS-COCO \cite{img_cap:ms_coco}. Then, they are divided into train ($36k$), test ($1k$), and validation ($5k$) sets.
For the image captioning task, we follow the split of \cite{img_cap:krishna}, which contains $113k$ images in training set, $5k$ images for testing, and $5k$ for validation, collected from MS-COCO \cite{img_cap:ms_coco}.

\noindent\textbf{Model Configuration}: As our Fine-Grained Predicates Learning (FGPL) and Adaptive Fine-Grained Predicates Learning (FGPL-A) are both model-agnostic, following recent works~\cite{prior:iccv}, we incorporate them into VCTree~\cite{ssg:vctree}, Motif~\cite{motifs}, and Transformer~\cite{networks:transformer} within the SGG benchmark~\cite{ssg:benchmark}.

\begin{table*}[ht]
\caption{\textbf{Comparison between existing methods and our methods (i.e., FGPL-A and FGPL) on three sub-tasks of mR@K(\%) on the VG-SGG dataset}. Reweight* denotes the re-implemented state-of-the-art re-weighting method proposed in~\cite{seesaw}.}
\centering
		\resizebox{0.85\textwidth}{!}{
\begin{tabular}{llccccccccc}
	\toprule
            \multirow{2}{*}{}&\multirow{2}{*}{Method}        &   \multicolumn{3}{c}{Predicate Classification (PredCls)} & \multicolumn{3}{c}{Scene Graph Classification (SGCls)} & \multicolumn{3}{c}{Scene Graph Detection (SGDet)}  \\ \cline{3-11}
                             &  & mR@20         & mR@50        & mR@100        & mR@20         & mR@50         & mR@100         & mR@20        & mR@50       & mR@100        \\ \hline
							    \multirow{9}{*}{\rotatebox{90}{Transformer}}&Transformer~\cite{networks:transformer}$_{\textit{NIPS'17}}$  & 12.4                  & 16.0          & 17.5            & 7.7           & 9.6          & 10.2           & 5.3           & 7.3        & 8.8   \\
				&-CogTree~\cite{cogtree}$_{\textit{IJCAI'21}}$ & 22.9          & 28.4          & 31.0          & 13.0            & 15.7           & 16.7          & 7.9           & 11.1           & 12.7          \\ 
				&-EQL~\cite{eql}$_{\textit{CVPR'20}}$ & 24.3          & 28.5            & 31.6            & 12.3           & 14.4          & 16.2           & 8.7           & 11.8        & 14.6  \\ 
				&-BASGG~\cite{prior:iccv}$_{\textit{ICCV'21}}$ & 26.7          & 31.9            & 34.2            & 15.7           & 18.5          & 19.4           & 11.4           & 14.8        & 17.1  \\  
				&-Reweight*~\cite{seesaw}$_{\textit{CVPR'21}}$ & 19.5                  & 28.6          & 34.4            & 11.9           & 17.2          & 20.7           & 8.1           & 11.5        & 14.9   \\
				&-HML~\cite{hml}$_{\textit{Arxiv'22}}$ & 27.5                  & 33.3          & 35.9            & 15.7           & 19.1          & 20.4           & 11.4           & 14.9        & 17.7   \\				
				&-IETrans~\cite{scenegraph:ietrans}$_{\textit{Arxiv'22}}$  & -          & 35.0          & 38.0          & -           & 20.8           & 22.3          & -           & 15.0           & 18.1           \\  
				&\textbf{-FGPL}~\cite{scenegraph:fgpl}$_{\textit{CVPR'22}}$           & \underline{27.5}         & \underline{36.4}        & \underline{40.3}         & \underline{19.2}         & \underline{22.6}         & \underline{24.0}          & \underline{13.2}         & \underline{17.4}       & \underline{20.3} \\ 
				&\textbf{-FGPL-A}     & \textbf{28.4}         & \textbf{38.0}        & \textbf{42.4}         & \textbf{20.5}         & \textbf{24.0}         & \textbf{25.4}          & \textbf{13.4}         & \textbf{18.0}       & \textbf{21.0}   \\ 
						\cline{2-11}
				\multirow{18}{*}{\rotatebox{90}{VCTree}} &VCTree$_{\textit{CVPR'19}}$  & 11.7          & 14.9          & 16.1          & 6.2           & 7.5           & 7.9          & 4.2           & 5.7           & 6.9           \\
		 				&-EBM~\cite{energy}$_{\textit{CVPR'21}}$  & 14.2          & 18.2            & 19.7            & 10.4           & 12.5          & 13.5           & 5.7           & 7.7        & 9.1  \\ 
		 				&-SG~\cite{sg}$_{\textit{ICCV'21}}$  & 15.0          & 19.2            & 21.1            & 9.3           & 11.6          & 12.3           & 6.3           & 8.1        & 9.0  \\ 
		 				&-PUM~\cite{prob}$_{\textit{CVPR'21}}$  & - & 20.2 & 22.0 & -  & 11.9  & 12.8 & - & 7.7 & 8.9 \\  
		 				 				&-NARE~\cite{scenegraph:nare}$_{\textit{CVPR'22}}$ & 18.0          & 21.7            & 23.1            & 11.9           & 14.1          & 15.2           & 7.1           & 8.2        & 8.7  \\  
 				 				 				&-PCPL~\cite{pcpl}$_{\textit{MM'21}}$  & -          & 22.8          & 24.5          & -           & 15.2           & 16.1          & -           & 10.8           & 12.6           \\ 
 				 				&-DLFE~\cite{scenegraph:dlfe}$_{\textit{MM'21}}$  & 20.8          & 25.3          & 27.1          & 15.8           & 18.9           & 20.0          & 8.6           & 11.8           & 13.8           \\
 				&-TDE~\cite{tde}$_{\textit{CVPR'20}}$   & 18.4          & 25.4          & 28.7          & 8.9            & 12.2           & 14.0          & 6.9           & 9.3           & 11.1          \\ 
 				&-CogTree~\cite{cogtree}$_{\textit{IJCAI'21}}$  & 22.0          & 27.6          & 29.7          & 15.4            & 18.8           & 19.9          & 7.8           & 10.4           & 12.1   \\
				&-BASGG~\cite{prior:iccv}$_{\textit{ICCV'21}}$ & 26.2          & 30.6            & 32.6            & 17.2           & 20.1          & 21.2           & 10.6           & 13.5        & 15.7  \\  

				&-NICE~\cite{NICE}$_{\textit{CVPR'22}}$ & -          & 30.7            & 33.0            & -           & 19.9          & 21.3           & -           & 11.9        & 14.1  \\   			   
			    &-PPDL~\cite{PPDL}$_{\textit{CVPR'22}}$ & -          & 33.3            & 33.8            & -           & 21.8          & 22.4           & -           & 11.3        & 13.3  \\  
			    &-Reweight*~\cite{seesaw}$_{\textit{CVPR'21}}$ & 19.4                  & 29.6          & 35.3            & 13.7           & 19.9          & 23.5           & 7.0           & 10.5        & 13.1   \\ 
			    &-RTPB~\cite{RTPB}$_{\textit{AAAI'22}}$ & 28.8          & 35.3            & 35.6            & 20.6           & 24.5          & 25.8           & 9.6           & 12.8        & 15.1  \\ 

				&-GCL~\cite{scenegraph:gcl}$_{\textit{CVPR'22}}$ & -          & 37.1          & 39.1          & -           & 22.5           & 23.5          & -           & 15.2           & 17.5           \\
				&-IETrans~\cite{scenegraph:ietrans}$_{\textit{Axiv'22}}$  & -          & 37.0          & 39.7          & -           & 19.9           & 21.8          & -           & 12.0           & 14.9           \\  
				&\textbf{-FGPL}~\cite{scenegraph:fgpl}$_{\textit{CVPR'22}}$           & \underline{30.8}         & \underline{37.5}        & \underline{40.2}         & \underline{21.9}         & \underline{26.2}         & \underline{27.6}          & \underline{11.9}         & \underline{16.2}       & \underline{19.1} \\ 
				&\textbf{-FGPL-A}          & \textbf{34.5}         & \textbf{41.6}        & \textbf{44.3}         & \textbf{24.1}         & \textbf{28.8}         & \textbf{30.6}          & \textbf{12.2}         & \textbf{16.9}       & \textbf{20.0} \\ 
			
			\cline{2-11}
				
				\multirow{18}{*}{\rotatebox{90}{Motif}}&Motif~\cite{motifs}$_{\textit{CVPR'18}}$  	& 11.5          & 14.6            & 15.8            & 6.5           & 8.0          & 8.5           & 4.1           & 5.5        & 6.8  \\
 				 &-Focal~\cite{tde}$_{\textit{CVPR'20}}$   & 10.9          & 13.9          & 15.0          & 6.3            & 7.7           & 8.3          & 3.9           & 5.3           & 6.6          \\
 				 &-EBM~\cite{energy}$_{\textit{CVPR'21}}$  & 14.2          & 18.0          & 19.5          & 8.2           & 10.2           & 11.0          & 5.7           & 7.7           & 9.3           \\ 
 				  	 &-Resample~\cite{tde}$_{\textit{CVPR'20}}$   & 14.7          & 18.5          & 20.0          & 9.1            & 11.0           & 11.8          & 5.9           & 8.2           & 9.7          \\ 				 
 				  &-SG~\cite{sg}$_{\textit{ICCV'21}}$  & 14.5          & 18.5            & 20.2            & 8.9           & 11.2          & 12.1           & 6.4           & 8.3        & 9.2  \\  	
 			
 				 &-C-bias~\cite{c-bias}$_{\textit{Axiv'22}}$  & 16.6          & 20.4            & 21.9           & 8.5          & 9.9           & 10.4           & 3.9        & 5.5 & 6.9  \\ 				
  				 &-PCPL~\cite{pcpl}$_{\textit{MM'20}}$  & -          & 24.3            & 26.1           & -          & 12.0           & 12.7           & -        & 10.7 & 12.6  \\				
 				 &-TDE~\cite{tde}$_{\textit{CVPR'20}}$  & 18.5          & 24.9            & 28.3           & 11.1          & 13.9           & 15.2           & 6.6        & 8.5 & 9.9  \\

 				 &-DLFE~\cite{scenegraph:dlfe}$_{\textit{MM'21}}$ & 22.1          & 26.9          & 28.8          & 12.8           & 15.2           & 15.9          & 8.6           & 11.7           & 13.8           \\ 
 							 &-CogTree~\cite{cogtree}$_{\textit{IJCAI'21}}$ & 20.9          & 26.4          & 29.0          & 12.1            & 14.9           & 16.1          & 7.9           & 10.4           & 11.8          \\ 
				&-BASGG~\cite{prior:iccv}$_{\textit{ICCV'21}}$ & 24.8          & 29.7            & 31.7            & 14.0           & 16.5          & 17.5           & 10.7           & 13.5        & 15.6  \\ 
			    &-NICE~\cite{NICE}$_{\textit{CVPR'22}}$ & -          & 29.9            & 32.3            & -           & 16.6          & 17.9           & -           & 12.2        & 14.4  \\  
				 &-PPDL~\cite{PPDL}$_{\textit{CVPR'22}}$ & -          & 32.2            & 33.3            & -           & 17.5          & 18.2           & -           & 11.4        & 13.5  \\  
				&-Reweight*~\cite{seesaw}$_{\textit{CVPR'21}}$ & 18.8                  & 28.1          & 33.7            & 10.7           & 15.6          & 18.3           & 7.2           & 10.5        & 13.2   \\ 
				&-GCL~\cite{scenegraph:gcl}$_{\textit{CVPR'22}}$  & -          & \underline{36.1}          & 38.2          & -           & 20.8           & 21.8          & -           & \underline{16.8}           & \underline{19.3}           \\ 
				&-IETrans~\cite{scenegraph:ietrans}$_{\textit{Axiv'22}}$  & -          & 35.8          & \underline{39.1}          & -           & \underline{21.5}           & \underline{22.8}          & -           & 15.5           & 18.0           \\  
				&\textbf{-FGPL}~\cite{scenegraph:fgpl}$_{\textit{CVPR'22}}$           & \underline{24.3}         & 33.0        & 37.5         & \underline{17.1}         & 21.3         & 22.5          & \underline{11.1}         & 15.4       & 18.2 \\ 
				&\textbf{-FGPL-A}         & \textbf{27.2}         & \textbf{36.3}        & \textbf{40.7}         & \textbf{19.9}         & \textbf{23.2}         & \textbf{24.5}          & \textbf{12.5}         & \textbf{17.0}       & \textbf{19.8}  \\

\bottomrule
\end{tabular}}

\label{tab.compare}
\end{table*}
	
\noindent\textbf{Evaluation Tasks}: Following recent works~\cite{ssg:vctree,bgnn}, we evaluate our methods on three sub-tasks of SGG, including PredCls (Predicate Classification), SGCls (Scene Graph Classification), and SGDet (Scene Graph Detection). In the PredCls, model needs to predict predicates given ground-truth object bounding boxes together with their labels. For the SGCls task, taking ground truth bounding boxes as input, the model predicts both object labels and predicates (relationships) between them. The SGDet task requires the SGG model to generate the object labels with relationships from scratch, i.e., without ground-truth bounding boxes and object labels. Moreover, following \cite{tde}, we conduct Sentence-to-Graph retrieval as a downstream task to verify the effectiveness of fine-grained scene graphs generated by our Adaptive Fine-Grained Predicate Learning (FGPL-A). Specifically, SGG models extract scene graphs under the SGDet setting (i.e., without annotated bounding boxes and object labels). Then, the generated scene graphs together with a corpus of image captions are embedded into a shared space, respectively. Finally, it queries scene graphs with the embedding of image captions on a gallery during inference. We report Recall@K on the 1k/5k gallery of image captions for our experiments. We also perform Image Captioning task to prove that the generated scene graphs precisely describe scenarios. Specifically, object visual representations and scene graphs are combined to generate image captions. Moreover, we evaluate the generated image captions with Bleu-4~\cite{img_cap:bleu}, Meteor~\cite{img_cap:meteor}, Cider~\cite{img_cap:cider}, and Spice~\cite{img_cap:spice}.

\begin{algorithm}[!htb]

\caption{Discriminatory Power (DP@K)}
\label{alg:AAA}
\begin{algorithmic}[1]

\REQUIRE Confusion Matrix $S' \in \mathbb{R}^{C \times C}$, with $s'_{i,j}\in[0,1]$. 

\ENSURE Models' Discriminatory Power among top-K hard-to-distinguish predicates, $ y^{K}$. 

\FOR {$i=0$ to $C$}
\STATE create a Set $\mathcal{V}'_i = topK(s'_{i,j|j \neq i})$ 

\FOR {$s'_{i,j}$ in $\mathcal{V}'_i$}

\STATE $ y^{K} \gets  y^{K} + (s'_{i,i} - s'_{i,j})/K$

\ENDFOR
\ENDFOR

\STATE $ y^{K} \gets  y^{K}/C$

\RETURN $y^{K}$

\end{algorithmic}
\end{algorithm}
\label{sec:dp}

\noindent\textbf{Evaluation Metrics}: Following recent works~\cite{ssg:vctree,bgnn}, we evaluate model's performance on mR@K/R@K and Group Mean Recall, i.e., Head, Body, and Tail. Besides, we introduce DP@K ($\%$) to indicate models' Discriminatory Power among top-k hard-to-distinguish predicates. Generally, DP@K is calculated by averaging the difference between the proportion of samples correctly predicted as $i$ and the proportion of samples misclassified as hard-to-distinguish predicates $j$ ($j \in \mathcal{V}'_i$). Furthermore, $\mathcal{V}'_i$ is defined as a set of top-k hard-to-distinguish predicates for predicate $i$. Especially, to figure out hard-to-distinguish predicates, we collect a normalized confusion matrix $S' \in \mathbb{R}^{C \times C}$ from the model's prediction results, with $s'_{i,j}$, which denotes the degree of confusion between the predicate pair $i$ and $j$. For each predicate category $i$, $k$ predicates with the highest $s'_{i,j}$ are chosen to construct $\mathcal{V}'_i$. In a word, a higher score of DP@K means more substantial discriminatory power among hard-to-distinguish predicates. Moreover, the intuitive explanation of DP@K with Peruse-code is shown in Alg.~\ref{alg:AAA}.

\begin{table*}[ht]
	\caption{\textbf{Quantitative results on the generalizability of CDL/EDL in FGPL and CDL-A/EDL-A in FGPL-A on the VG-SGG dataset.} We validate the generalization capability of our proposed components, i.e., Entity Discriminating Loss (EDL) and Category Discriminating Loss (CDL), Adaptive Entity Discriminating Loss (EDL-A), and Adaptive Category Discriminating Loss (CDL-A) on the VG-SGG dataset in comparison to benchmark models.}
	\centering
		\resizebox{0.9\textwidth}{!}{
\begin{tabular}{lccccccccc}
	\toprule
                        \multirow{2}{*}{Method}    &         \multicolumn{3}{c}{Predicate Classification (PredCls)} & \multicolumn{3}{c}{Scene Graph Classification (SGCls)} & \multicolumn{3}{c}{Scene Graph Detection (SGDet)}  \\ \cline{2-10}
                  & mR@20         & mR@50        & mR@100        & mR@20         & mR@50         & mR@100         & mR@20        & mR@50       & mR@100        \\ \hline
            Transformer                & 12.4          & 16.0         & 17.5          & 7.7           & 9.6           & 10.2           & 5.3          & 7.3         & 8.8           \\
            ~+\textit{CDL}      & 23.0 $\uparrow$ \textbf{10.6}        & 31.4 $\uparrow$ \textbf{15.4}       & 35.4 $\uparrow$ \textbf{17.9}         & 14.3 $\uparrow$ \textbf{6.6}          & 18.9 $\uparrow$ \textbf{9.3}        & 21.2 $\uparrow$ \textbf{11.0}              & 9.4 $\uparrow$ \textbf{4.1}        & 13.3 $\uparrow$ \textbf{6.0}       & 16.5 $\uparrow$ \textbf{7.7}       \\
            ~+\textit{CDL+EDL}      & 27.5 $\uparrow$ \textbf{15.1}        & 36.4 $\uparrow$ \textbf{20.4}       & 40.3 $\uparrow$ \textbf{22.8}         & 19.2 $\uparrow$ \textbf{11.5}          & 22.6 $\uparrow$ \textbf{13.0}        & 24.0 $\uparrow$ \textbf{13.8}              & 13.2 $\uparrow$ \textbf{7.9}        & 17.4 $\uparrow$ \textbf{10.1}       & 20.3 $\uparrow$ \textbf{11.5}       \\            
            ~+\textit{CDL-A}      & 26.4 $\uparrow$ \textbf{14.0}        & 35.1 $\uparrow$ \textbf{19.1}       & 38.9 $\uparrow$ \textbf{21.4}         & $ 14.9\uparrow$ \textbf{7.2}          & 19.9 $\uparrow$ \textbf{10.3}        & 22.9 $\uparrow$ \textbf{12.7}              & 11.6 $\uparrow$ \textbf{6.3}        & 16.1 $\uparrow$ \textbf{8.8}       & 19.3 $\uparrow$ \textbf{10.5}       \\
                     
            ~+\textit{CDL-A+EDL-A}      & 28.4 $\uparrow$ \textbf{16.0}        & 38.0 $\uparrow$ \textbf{22.0}       & 42.4 $\uparrow$ \textbf{24.9}         & 20.5 $\uparrow$ \textbf{12.8}          & 24.0 $\uparrow$ \textbf{14.4}        & 25.4 $\uparrow$ \textbf{15.2}              & 13.4 $\uparrow$ \textbf{8.1}        &  18.0 $\uparrow$ \textbf{10.7}       & 21.0 $\uparrow$ \textbf{12.2}       \\ \hline
            
            VCTree                     & 11.7          & 14.9         & 16.1          & 6.2           & 7.5           & 7.9            & 4.2          & 5.7         & 6.9           \\
            ~+CDL           & 23.0 $\uparrow$ \textbf{11.3}         & 31.6  $\uparrow$ \textbf{16.7}       & 35.3 $\uparrow$ \textbf{19.2}         & 15.7 $\uparrow$ \textbf{9.5}          & 21.1 $\uparrow$ \textbf{13.6}         & 23.3 $\uparrow$ \textbf{15.4}   & 11.0 $\uparrow$ \textbf{6.8}       & 14.7 $\uparrow$ \textbf{9.0}       & 17.5 $\uparrow$ \textbf{10.6}        \\
             ~+\textit{CDL+EDL}      & 30.8 $\uparrow$ \textbf{19.1}        & 37.5 $\uparrow$ \textbf{22.6}       & 40.2 $\uparrow$ \textbf{24.1}         & 21.9 $\uparrow$ \textbf{15.7}          & 26.2 $\uparrow$ \textbf{18.7}        & 27.6  $ \uparrow$ \textbf{19.7}              & 11.9 $\uparrow$ \textbf{7.7}        & 16.2 $\uparrow$ \textbf{10.5}       & 19.1 $\uparrow$ \textbf{12.2}       \\           
            ~+\textit{CDL-A}      & 24.7 $\uparrow$ \textbf{13.0}        & 33.6 $\uparrow$ \textbf{18.7}       & 37.8 $\uparrow$ \textbf{21.7}         & 17.9 $\uparrow$ \textbf{11.7}          & 23.7 $\uparrow$ \textbf{16.2}        & 26.4 $\uparrow$ \textbf{18.5}              & 11.1 $\uparrow$ \textbf{6.9}        & 15.0 $\uparrow$ \textbf{9.3}       & 17.8 $\uparrow$ \textbf{10.9}       \\
           
            ~+\textit{CDL-A+EDL-A}      & 34.5 $\uparrow$ \textbf{22.8}        & 41.6 $\uparrow$ \textbf{26.7}       & 44.3 $\uparrow$ \textbf{28.2}         & 24.1 $\uparrow$ \textbf{17.9}          & 28.8 $\uparrow$ \textbf{21.3}        & 30.6 $\uparrow$ \textbf{22.7}              & 12.2 $\uparrow$ \textbf{8.0}        &  16.9 $\uparrow$ \textbf{11.2}       & 20.0 $\uparrow$ \textbf{13.1}       \\ \hline
            
            Motif             & 11.5          & 14.6         & 15.8          & 6.5           & 8.0           & 8.5            & 4.1          & 5.5         & 6.8           \\
            ~+\textit{CDL}         & 22.2 $\uparrow$ \textbf{10.7}         & 30.3 $\uparrow$ \textbf{15.7}        & 34.4 $\uparrow$ \textbf{18.6}          & 12.6 $\uparrow$ \textbf{6.1}          & 16.7 $\uparrow$ \textbf{8.7}        & 18.5 $\uparrow$ \textbf{10.0}         & 8.2 $\uparrow$ \textbf{4.1}        & 11.6 $\uparrow$ \textbf{6.1}      & 14.3 $\uparrow$ \textbf{7.5}        \\
            ~+\textit{CDL+EDL}      & 24.3 $\uparrow$ \textbf{12.8}        & 33.0 $\uparrow$ \textbf{18.4}       & 37.5 $\uparrow$ \textbf{21.7}         & 17.1 $\uparrow$ \textbf{10.6}          & 21.3 $\uparrow$ \textbf{13.3}        & 22.5 $ \uparrow$ \textbf{14.0}              & 11.1 $\uparrow$ \textbf{7.0}        & 15.4 $\uparrow$ \textbf{9.9}       &  18.2 $\uparrow$ \textbf{11.4}       \\     
            ~+\textit{CDL-A}      &  24.3 $\uparrow$ \textbf{12.8}        & 33.5 $\uparrow$ \textbf{18.9}       & 37.9 $\uparrow$ \textbf{22.1}         & 13.9 $\uparrow$ \textbf{7.4}          & 18.4 $\uparrow$ \textbf{10.4}        & 20.4 $\uparrow$ \textbf{11.9}              & 8.4 $\uparrow$ \textbf{4.3}        & 12.2 $\uparrow$ \textbf{6.7}       & 15.1 $ \uparrow$ \textbf{8.3}       \\
                   
            ~+\textit{CDL-A+EDL-A}      & 27.2 $\uparrow$ \textbf{15.7}        & 36.3 $\uparrow$ \textbf{21.7}       & 40.7 $\uparrow$ \textbf{24.9}         & 19.9 $\uparrow$ \textbf{13.4}          & 23.2 $\uparrow$ \textbf{15.2}        & 24.5 $\uparrow$ \textbf{16.0}              & 12.5 $\uparrow$ \textbf{8.4}        & 17.0 $\uparrow$ \textbf{11.5}       & 19.8 $\uparrow$ \textbf{13.0}       \\

          \bottomrule
\end{tabular}
}

	\label{tab.generation1}
\end{table*}

\subsection{Implementation Details}
\noindent\textbf{Detector}: For object detectors, we utilize the pre-trained Faster R-CNN by~\cite{obj_det:faster,tde} to detect objects in images. Moreover, object detectors' weights are frozen during scene graph generation training for all three sub-tasks. 

\noindent\textbf{Scene Graph Generation Model}: Following~\cite{ssg:benchmark}, benchmark models in Model Zoo are all trained with Cross-Entropy Loss and SGD optimizer with an initial learning rate of $0.01$, batch size as $16$. For the GQA-SGG dataset, we follow the standard Faster R-CNN settings to pre-train our object detector.

\noindent\textbf{Fine-Grained Predicates Learning}: We incorporate our Fined-Grained Predicate learning framework (FGPL) into benchmark models in Model Zoo~\cite{ssg:benchmark}. Moreover, they share the same hyper-parameters for Category Discriminating Loss (CDL) and Entity Discriminating Loss (EDL). In particular, we set $\alpha$, $\beta$, and $\xi$ as $1.5$, $2.0$, and $0.9$ for CDL. Additionally, based on the correlations derived from Predicate Lattice, we set the number of hard-to-distinguish predicates (i.e., $\left| {\mathcal{V}_i } \right|$) as $5$ for EDL. Furthermore, the boundary margin $\delta$ between hard-to-distinguish predicate pairs is uniformly set to be 0.5. Finally, we empirically set $\lambda$ as $0.1$.

\noindent\textbf{Adaptive Fine-Grained Predicate Learning}: Similar to FGPL, we integrate our FGPL-A into three benchmark SGG models (i.e., Motif, Transformer, and VCTree). Except for the refinement mechanism, the construction procedure of Adaptive Predicate Lattice is the same as that of FGPL. Moreover, the hyper-parameters of Adaptive Category/Entity Discriminating Loss (CDL-A/EDL-A) in FGPL-A share the same values with Category/Entity Discriminating Loss (CDL/EDL) in FGPL. Furthermore, the coefficient of the Batch-Refinement (BR) regime, $\tau$ is set as $0.99$. Finally, the hyper-parameter $\theta$ in CDL-A is set as $0.1$. 
Besides, for GQA-SGG datatset, parameters for FGPL-A and FGPL are set the same as those in the VG-SGG datatset. Due to the limitation of GPU's memory, we only conduct experiments on two benchmark models, i.e., Motif and Transformer, for the GQA-SGG dataset. 

\noindent\textbf{Sentence to Graph Retrieval:} Following \cite{tde}, we conduct the Sentence-to-Graph Retrieval task to validate the effectiveness of fine-grained scene graphs generated by our FGPL-A. The core of the task is to establish matching between semantic graphs of image captions and scene graphs. Specifically, semantic graphs are decomposed from image captions utilizing an off-the-shelf toolkit \cite{sen2graph:gen}. Then, the attention mechanism both embeds scene graphs and semantic graphs into the feature space, optimized by the contrastive loss. 

\noindent\textbf{Image Captioning:} We implement the image captioning task based on \cite{img_cap:ruotian} and select the Transformer \cite{cap:benchmark} as the baseline. Specifically, each scene graph is divided into several ``subject($o_i$)-predicate($r_{i,j}$)-object($o_j$)'' triplets. Then, for each image, we choose the most confident triplet with the highest score.
Furthermore, the self-attention mechanism encodes scene graphs and queries the instance visual features to predict the next word in the decoding process. 

\noindent\textbf{Experimental Devices:} We perform all the experiments on the server with Ubuntu 20.04.4 LTS and 4 NVIDIA GeForce RTX 3090 GPUs. Our codes are implemented with PyTorch 1.9.0.

\begin{table}[t]
	\caption{\textbf{Quantitative results on the generalizability of FGPL and FGPL-A on the GQA-SGG dataset.} We validate the generalization capability of our FGPL and FGPL-A on GQA-SGG in comparison to benchmark SGG models.}
	\centering
		\resizebox{0.98\linewidth}{!}{
            \begin{tabular}{lcccc}
	            \toprule	
            \multirow{2}{*}{Method} & \multicolumn{1}{c}{PredCls}& \multicolumn{1}{c}{SGCls} & \multicolumn{1}{c}{SGDet}  \\\cline{2-5}
                                      & mR@100/R@100         & mR@100/R@100         & mR@100/R@100      \\ \hline
             Motif               & 4.6/58.6      & 2.1/\textbf{24.1}      & 1.9/23.3                       \\
            ~-FGPL         &6.1/\textbf{58.8}         & 2.7/23.0     & 2.3/\textbf{23.8}        \\
            ~-FGPL-A         &\textbf{7.7}/57.7         & \textbf{3.1}/23.0       &\textbf{2.9}/23.3         \\ \hline
   
            Transformer         & 4.5/58.7          & 2.3/\textbf{24.1}         &   1.7/23.3                  \\
            ~-FGPL         &5.2/57.3         & 3.0/23.3     &   3.1/23.4     \\
            ~-FGPL-A         &\textbf{7.9/58.7}         & \textbf{3.5}/23.5     &\textbf{3.2/23.4}         &\textbf{}\\
            
          \bottomrule
\end{tabular}}

	\label{tab.gqa}
\end{table}

\subsection{Comparison with State of the Arts}
We compare our methods with state-of-the-art SGG models via incorporating FGPL and FGPL-A into three SGG benchmark models, namely Transformer~\cite{networks:transformer}, Motif~\cite{motifs}, and VCTree~\cite{ssg:vctree}.
Quantitative results compared with state-of-the-art methods on the VG-SGG dataset are shown in Tab.~\ref{tab.compare}.
Firstly, VCTree (FGPL-A), Transformer (FGPL-A), and Motif (FGPL-A) outperform all the state-of-the-art methods on all SGG tasks of all the metrics, achieving $44.3\%$, $42.4\%$ and $40.7\%$ of VCTree (FGPL-A), Transformer (FGPL-A) and Motif (FGPL-A) on mR@100 under the PredCls task.
Specifically, compared with the state-of-the-art SGG method IETrans \cite{scenegraph:ietrans}, FGPL-A achieves superior performance on three benchmark models of all the metrics, which demonstrates the effectiveness of FGPL-A on predicates discrimination.  
Besides, our FGP~\cite{scenegraph:fgpl} also achieves surpassing or comparable performance in comparison to previous methods.
Finally, to verify the significant efficacy of predicate correlations for improving discriminatory power among predicates, we make comparisons between benchmark models trained with the Re-weight and FGPL-A learning strategies. We observe that compared with Reweight*-Motif, Reweight*-VCTree, and Reweight*-Transformer, FGPL-A-Motif, FGPL-A-VCTree, and FGPL-A-Transformer achieve large margins of improvements by $7.0\%$, $9.0\%$, and $8.0\%$ on mR@100 for PredCls, verifying the effectiveness of FGPL-A. 
Intuitively, fully understanding relationships among predicates, our method can adjust the re-weighting process based on predicate correlations, which boosts and sustains the discriminatory capability over hard-to-distinguish and recognizable predicates, respectively.

\subsection{Generalization on SGG Models}
In this section, we verify the generalizability of FGPL and FGPL-A on different SGG benchmark models under two SGG datasets, i.e., VG-SGG and GQA-SGG. 

\noindent\textbf{Quantitative analysis on VG-SGG:}
To verify the CDL, EDL of FGPL and the CDL-A, EDL-A within FGPL-A are all plug-and-play, we gradually incorporate them into different benchmark models (i.e., Transformer, VCTree and Motif) on the VG-SGG dataset.
Quantitative results are shown in Tab.~\ref{tab.generation1}. 
Integrated with CDL, we observe considerate improvements as at least $17.9\%$ on three benchmark models of mR@100 under the PredCls task, showing the notable generalizability for CDL on figuring out and differentiating among hard-to-distinguish predicates.
Besides, after EDL is applied to benchmark models, the performance further boosts about $5.0\%$ on mR@100 under the PredCls task, which manifests the remarkable compatibility of EDL. The possible reason is that EDL adjusts the learning process according to contexts varied with the training samples.
Moreover, after being extended from CDL to CDL-A, Transformer (CDL-A), VCTree (CDL-A), and Motif (CDL-A) achieve more considerable gains as $3.5\%$, $2.5\%$, and $3.5\%$ on mR@100 under the PredCls task than Transformer (CDL), VCTree (CDL), and Motif (CDL). It demonstrates the effectiveness in two aspects: $1$) The progressively assigned targets increase model's discriminatory power. $2$) The softened targets provide SGG models with comprehensive understanding for fine-grained predicate discrimination.
Finally, Transformer (CDL-A+EDL-A), VCTree (CDL-A+EDL-A), and Motif (CDL-A+EDL-A) achieve the best performance against benchmark models, indicating the generalizability of both CDL-A and EDL-A. 
We conjecture that by exploring the refined predicate correlations during training, both CDL-A and EDL-A make model's learning procedure consistent with its gradually gained discriminatory power, ensuring a robust and efficient learning process.  

\noindent\textbf{Quantitative analysis on GQA-SGG:}
We further verify the generalizability of our FGPL and FGPL-A on a more challenging dataset, namely GQA-SGG. Moreover, quantitative results are shown in Tab.~\ref{tab.gqa}.
From Tab.~\ref{tab.gqa}, benchmark models integrated with FGPL-A and FGPL outperform benchmark models by large margins on mR@100, and meanwhile achieve the comparable performance on R@100 on three SGG sub-tasks (e.g., Transformer (FGPL-A) \textit{vs.} Transformer (FGPL) \textit{vs.} Transformer: $7.9\%/58.7\%$ \textit{vs.}~$5.2\%/57.3\%$ \textit{vs.}~$4.5\%/58.7\%$, on mR@100/R@100 under the PredCls task).
It verifies FGPL and FGPL-A's generalizability on boosting model's discriminatory power among hard-to-distinguish predicates while maintaining that on recognizable categories.

\subsection{Predicate Discrimination Analysis of FGPL and FGPL-A}
\label{sec.sdl}
To certify the assumption that FGPL and FGPL-A strengthen models' discriminatory power among hard-to-distinguish predicates while maintaining a balanced learning process, we give quantitative and qualitative analysis of Predicate Discriminatory Power in Sec.~\ref{PDP} and Balanced Predicate Discrimination in Sec.~\ref{BPD} of FGPL and FGPL-A on the VG-SGG dataset.
\begin{figure}[t]
\begin{center}
\includegraphics[width=0.37\textwidth]{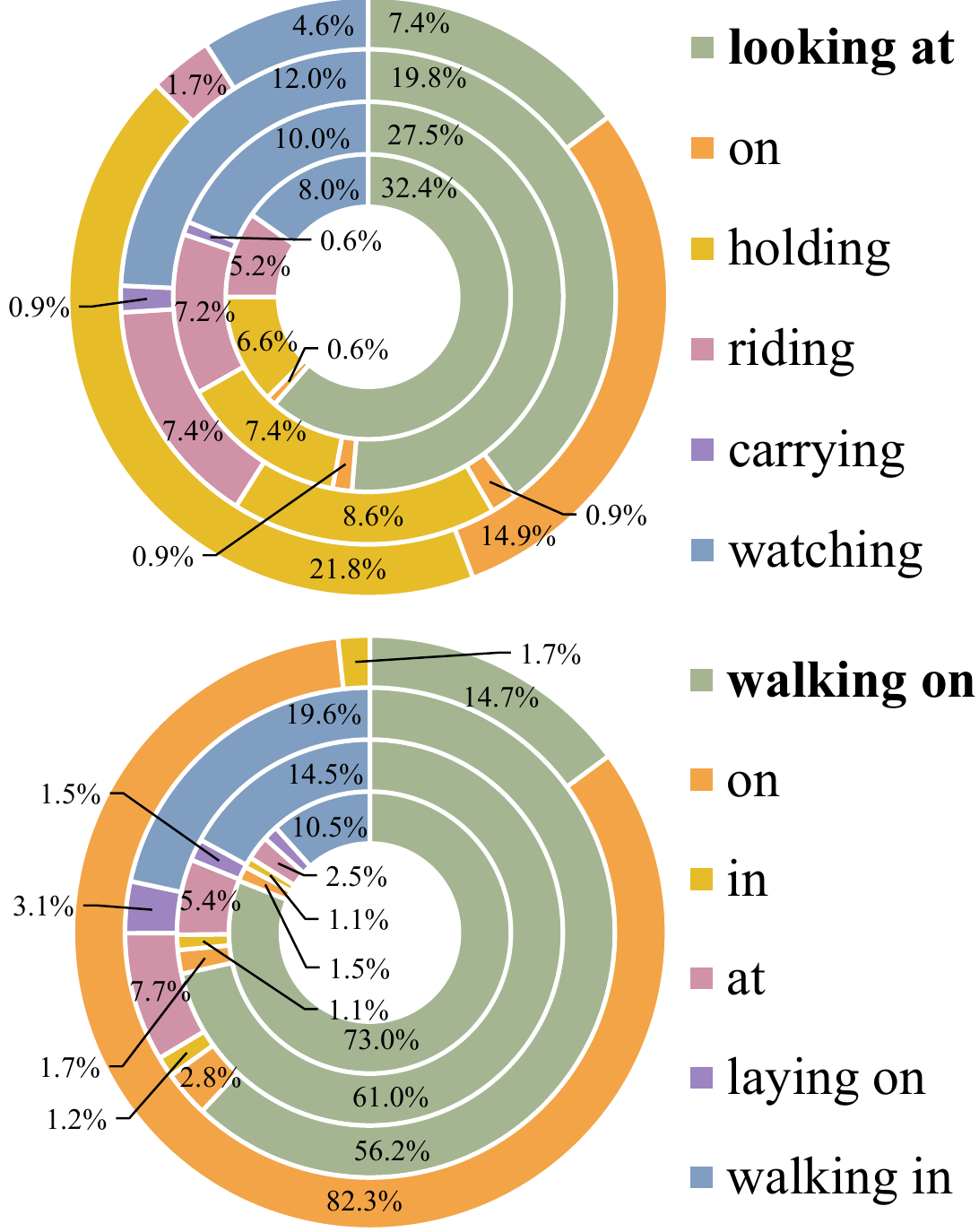}
\caption{\textbf{Effectiveness of FGPL and FGPL-A among hard-to-distinguish predicates.} Inner-ring (Transformer(FGPL-A)), first-middle-ring (Transformer (FGPL)), second-middle-ring (Transformer (Re-weight)), and outer-ring (Transformer) represent prediction distribution of hard-to-distinguish predicates acquired from different models, for samples with ground truth as \textbf{``looking at''} above, \textbf{``walking on''} below on the VG-SGG dataset.}
\label{fig:ring}
\end{center}

\end{figure}
\begin{table}[t]
	\caption{\textbf{Quantitative results on discriminatory power among top-k hard-to-distinguish predicates (DP@K(\%)) under the PredCls task on the VG-SGG dataset.} Re-weight denotes the re-implemented state-of-the-art re-weighting method proposed in~\cite{seesaw}.}
	\centering
		\resizebox{0.8\linewidth}{!}{
\begin{tabular}{lcccc}
	\toprule
            \multirow{2}{*}{Method}    &     \multicolumn{4}{c}{Predicate Classification (PredCls)}  \\ \cline{2-5}
                                      & DP@5          & DP@10         & DP@20  & Mean    \\ \hline
            Transformer             & 15.6          & 17.4          & 18.5      & 17.2           \\
            ~-Re-weight             & 33.3          & 36.1          & 38.1       &35.8\\
            ~-FGPL                    &37.9           & 40.3          & 42.1      & 40.1  \\
            ~-FGPL-A             &\textbf{38.6}         & \textbf{41.1}       &\textbf{42.9}    & \textbf{40.3}    \\ \hline
            VCTree                 & 14.1          & 15.7         & 17.3         & 15.7    \\
            ~-Re-weight              & 33.9          & 36.5          & 38.4      & 36.3 \\
            ~-FGPL                  &35.4         &37.8       &39.6               & 37.6    \\ 
            ~-FGPL-A                 &\textbf{36.6}        & \textbf{39.3}     &\textbf{41.1}   &  \textbf{39.0}     \\
            \hline
            Motif                 & 15.4          & 17.1          & 18.2          &   16.9       \\
            ~-Re-weight             & 33.1          &  35.7         & 37.5          &  35.4           \\
            ~-FGPL                 & 36.1          & 38.7         & 40.6        &  38.5       \\
            ~-FGPL-A               &\textbf{37.1}         & \textbf{39.6}     &\textbf{41.5} & \textbf{39.4}\\
           \bottomrule
\end{tabular}}

	\label{tab.hard-to-distinguishness}
\end{table}
\subsubsection{Analysis of Predicate Discriminatory Power}
\label{PDP}

\begin{figure*}[t]
  \centering
  \includegraphics[width=\textwidth]{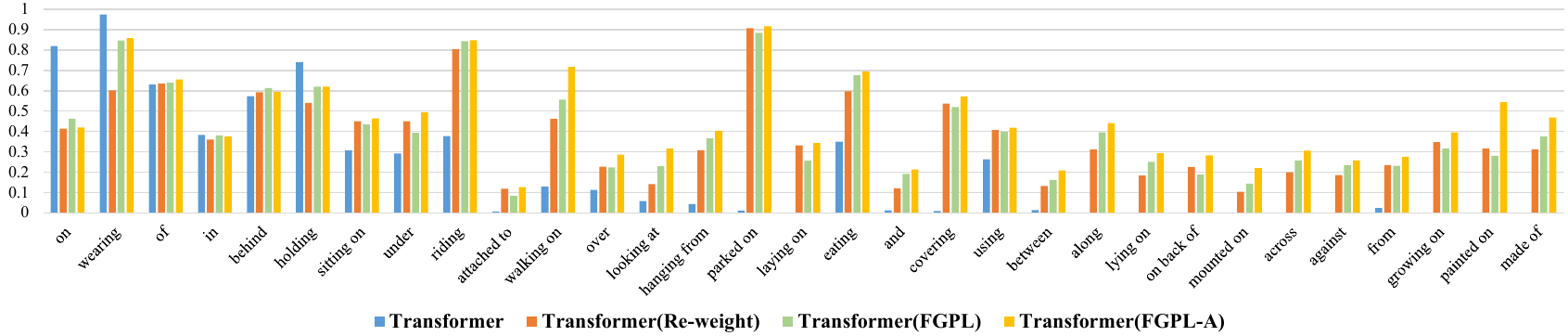}
  \setlength{\abovecaptionskip}{0.cm}
        \caption{\textbf{R@100 for predicates under PredCls task among Transformer, Transformer (Re-weight), Transformer (FGPL), and Transformer (FGPL-A).} Transformer (FGPL) and Transformer (FGPL-A) achieve more balanced and effective predicate discrimination among predicates with different frequencies than Transformer and Transformer (Re-weight).}
  \label{fig:recall2}
\end{figure*}
\noindent\textbf{Quantitative Analysis:} As hypothesized, our FGPL and FGPL-A improve model's discriminatory power among hard-to-distinguish predicates compared with benchmarks SGG models and the re-weighting method. Accordingly, we conduct experiments among four settings on the VG-SGG dataset to testify our hypothesis: 1) Benchmark models with vanilla Cross-Entropy Loss. 2) Benchmark models with state-of-the-art re-weighting method in~\cite{seesaw}. 3) Benchmark models with our FGPL. 4) Benchmark models incorporated with FGPL-A.
Tab.~\ref{tab.hard-to-distinguishness} presents comparisons among Transformer, VCTree, and Motif under PredCls task evaluated by DP@K which reveals model's discriminatory capability among hard-to-distinguish predicates. After being integrated with FGPL, Transformer (FGPL), VCTree (FGPL), and Motif (FGPL) greatly surpass their benchmark models on DP@10 with a large margin as $22.9\%$ , $22.1\%$, and $22.1\%$. 
It provides direct evidence that our FGPL considerably enhances model's discriminatory power among hard-to-distinguish predicates. It is also important to note that Transformer (FGPL), VCTree (FGPL), and Motif (FGPL) achieve consistent progress on DP@10 compared with Transformer (Re-weight), VCTree (Re-weight), and Motif (Re-weight). It reflects that our FGPL improves discriminatory ability over re-weighting method~\cite{seesaw} to generate fine-grained predicates. 
The possible reason is that FGPL makes the learning process both adapt to correlations of predicates and inherent contextual information of each sample, strengthening model's capability of differentiating hard-to-distinguish predicates.
After extending FGPL to FGPL-A, we observe a further boost as $0.8\%$, $1.5\%$, and $0.9\%$ on DP@10 in comparison to Transformer (FGPL), VCTree (FGPL), and Motif (FGPL). The compelling results powerfully manifest that adaptively refining the predicate correlations in keeping with model's learning status greatly benefits its discriminatory ability in differentiating hard-to-distinguish predicates.

\noindent\textbf{Qualitative Analysis:} For an intuitive illustration of FGPL and FGPL-A's discriminatory power, we visualize model's prediction distribution of hard-to-distinguish predicates from Transformer, Transformer (Re-weight), Transformer (FGPL), and Transformer (FGPL-A) on the VG-SGG dataset, shown in Fig.~\ref{fig:ring}. The proportion of rings indicates the distribution of prediction results, including hard-to-distinguish predicates $j$ and ground truth predicates $i$, for all samples with ground truth $i$. For predicate ``looking at'' in Fig.~\ref{fig:ring}, Transformer struggles to distinguish it from its correlated predicates, e.g., ``on'' or ``holding''. Besides, Transformer (Re-weight) fails to distinct ``looking at'' from hard-to-distinguish predicates, e.g., ``holding'', ``carrying'', and ``riding''. For Transformer (FGPL), the proportion of correctly classified samples rises from $7.4\%$ to $27.5\%$ compared with Transformer. Meanwhile, hard-to-distinguish predicates are more recognizable than Transformer (Re-weight), i.e., ``holding'' dropping from $8.6\%$ to $7.4\%$ and ``watching'' from $12.0\%$ to $10.0\%$. Consequently, the results validate our FGPL's efficiency of discriminatory capability among hard-to-distinguish predicates.  
Furthermore, after being upgraded from FGPL to FGPL-A, it yields a remarkable enhancement on mode's discriminatory power. For instance, for Transformer (FGPL-A), the proportion of correctly predicted samples rises from $27.5\%$ to $32.4\%$, while the negative categories ``holding'' drops from $7.4\%$ to $6.6\%$ and ``riding'' from $7.2\%$ to $5.2\%$ in comparison to Transformer (FGPL). The results substantiate that dynamically selecting hard-to-distinguish predicates and progressively setting fine-grained learning targets improve model's discriminatory power in differentiating among hard-to-distinguish predicates.


\subsubsection{Analysis of Balanced Predicate Discrimination}
\label{BPD}
\begin{table}[t]
\caption{\textbf{Quantitative comparisons among benchmark models integrated with different learning strategies (Re-weight, FGPL and FGPL-A) on Group Mean Recall@100 of the VG-SGG dataset.} We evaluate the performance of benchmarks trained with FGPL and FGPL-A in comparison to the Re-weight method on the Group Mean Recall under the PredCls task.}
	\centering
		\resizebox{0.9\linewidth}{!}{
\begin{tabular}{lcccc}
	\toprule
            \multirow{2}{*}{Method}    &     \multicolumn{4}{c}{Predicate Classification (PredCls)}  \\ \cline{2-5}
                                & Head (17)       & Body (17)         & Tail (16) &Mean  \\ \hline
            Transformer         & 38.8            & 9.6        & 3.1      & 17.2\\
            ~-Re-weight         & 39.8            & 34.2       & 28.8     &34.3\\
            ~-FGPL              & 42.2            & 37.8       & 40.7     &40.2\\
            ~-FGPL-A            & \textbf{42.2}   &\textbf{42.7}     &\textbf{42.4}     & \textbf{42.4}\\ \hline
            VCTree              & 39.5   & 9.1         & 2.9   &17.2\\
            ~-Re-weight         & 40.9            & 36.1        & 28.6   &35.2\\
            ~-FGPL              & 38.4            & \textbf{43.7}        & 38.3   &40.1\\
            ~-FGPL-A            & \textbf{43.2}  &39.8           & \textbf{39.7}  &\textbf{40.9}\\ \hline
            Motif               & 40.0            & 9.8      & 2.5          &17.4\\
            ~-Re-weight         & 40.7            & 35.1      & 24.7   &33.5\\
            ~-FGPL              & \textbf{43.4}   & 35.8  & 33.5 &37.6\\
            ~-FGPL-A            & 41.9   & \textbf{40.5}  & \textbf{39.8} &\textbf{40.7}\\
                    \bottomrule
\end{tabular}
}

	\label{tab.2}
\end{table}

\noindent\textbf{Quantitative Analysis:}
To verify that FGPL and FGPL-A guarantee the balanced predicate discrimination among predicates with different frequencies, we make comparisons among benchmark models (i.e., Transformer, VCTree, and Motif) incorporated with different learning strategies (i.e., Re-weight, FGPL and FGPL-A) on Group Mean Recall metrics under the PredCls task. Particularly, 50 categories of predicates are sorted and divided into Head Group (17), Body Group (17), and Tail Group (16) according to their frequency within the VG-SGG dataset.
The experimental results are shown in Tab.~\ref{tab.2}. 
Generally, after being integrated with FPGL or FPGL-A, we observe significant enhancements on Group Mean Recall metrics.
Specifically, compared with Transformer and Transformer (Re-weight), both Transformer (FGPL-A) and Transformer (FGPL) achieve more balanced performance among different predicate groups (e.g., Transformer (FGPL-A): Head-$42.2\%$, Body-$42.7\%$, Tail-$42.4\%$, Mean-$42.4\%$. Transformer (FGPL): Head-$42.2\%$, Body-$37.8\%$, Tail-$40.7\%$, Mean-$40.2\%$), which validates FGPL-A and FGPL's ability to alleviate imbalanced learning issues of the general and Re-weighting SGG methods.
It is worth to note that, although Motif (FGPL-A) fails to outperform Motif (FGPL) on the Head Group (i.e., $41.9\%$ \textit{vs.}~$43.4\%$), Motif (FGPL-A) still achieves a more balanced discrimination with a minor difference among different predicate groups than Motif (FGPL).
It demonstrates FGPL-A's effectiveness in guaranteeing a balanced learning procedure, which adjusts the training process in keeping with model's learning status.


\begin{table}[]
	\caption{\textbf{Ablation study on PC and RF of CDL}. PC denotes Predicate Correlation. RF denotes the Re-weighting Factor. The results are obtained with Transformer as the baseline.}
	\centering
	\resizebox{0.8\linewidth}{!}{
	\begin{tabular}{cc|cccc}
	\toprule
		\multicolumn{2}{c}{CDL}                       & \multicolumn{4}{c}{Predicate Classification (PredCls)} \\ \hline
		PC              & RF                             & mR@100   & DP@5 & DP@10 & DP@20 \\ \hline
		$\times$        & $\times$                       & 17.5         & 15.6        & 17.4      & 18.5\\
		$\times$        & $\checkmark$                   & 34.4         & 33.3        & 36.1      & 38.1\\
		$\checkmark$    & $\checkmark$                   & \textbf{40.3}         & \textbf{37.9}        & \textbf{40.3}      & \textbf{42.1}      \\  \bottomrule
	\end{tabular}}

	\label{tab:ablationCA_1}
\end{table}

\noindent\textbf{Qualitative Analysis:} 
For an intuitive illustration of FGPL and FGPL-A's balanced predicate discrimination, we provide Recall@100 results on predicates from Transformer, Transformer (Re-weight), Transformer (FGPL), and Transformer (FGPL-A), as shown in Fig.~\ref{fig:recall2}. For clarify, we choose 30 predicates with the highest frequency from each predicate group.
We observe that Transformer (FGPL) outperforms Transformer and Transformer (Re-weight) on almost all the predicates, demonstrating its significant efficacy of ensuring a more effective and balanced learning process among different categories. 
After being incorporated with FGPL-A, Transformer (FGPL-A) gains further progress on most of the predicate classes, which manifests the effectiveness of FGPL-A in guaranteeing an efficient learning process.
The possible explanation is that FGPL-A dynamically determines whether predicates pairs are hard-to-distinguish or not during training, which enhances model's discriminatory power among hard-to-distinguish predicates and meanwhile preserves that on recognizable ones, resulting in a robust and efficient learning procedure. 
Although Transformer (FGPL) and Transformer (FGPL-A) achieve superior performance compared with Transformer, we observe a reduction of performance on some predicates from the Head Group, e.g., ``on'', ``wearing'' and ``holding''.
The drops of Recall on those head predicates are inevitable in fine-grained classification as observed in long-tailed tasks~\cite{LVIS}. As claimed in~\cite{prior:iccv}, the general SGG models are over-confident on head classes with a high Recall.
For discriminating among hard-to-distinguish predicates, FGPL and FGPL-A classify some head classes (e.g., ``on'') into their fine-grained ones of tail classes (e.g., ``standing on''). 
It is prone to cause a degradation in head classes' performance, but meanwhile improves models' discriminatory power on tail ones with higher performance on DP@K and mR@K, as shown in Tab.~\ref{tab.hard-to-distinguishness} and Tab.~\ref{tab.compare}.

\subsection{Ablation Study}
To deeply investigate our FGPL and FGPL-A, we further study different ablation variants of CDL/CDL-A, EDL/EDL-A, and PL-A under the PredCls task on the VG-SGG dataset.

\begin{table}[]
	\caption{\textbf{Ablation study on PC and BF of EDL}. PC and BF denote Predicate Correlation and Balancing Factor, respectively. The results are obtained with Transformer as the baseline.}
	\centering
	\resizebox{0.8\linewidth}{!}{
	\begin{tabular}{cc|cccc}
	\toprule 
		\multicolumn{2}{c}{EDL}     & \multicolumn{4}{c}{Predicate Classification (PredCls)} \\ \hline
		PC  & BF   & mR@100   & DP@5 & DP@10 & DP@20 \\ \hline
		$\times$             & $\times$               & 17.7         & 15.9        & 17.7      & 18.8      \\
		$\checkmark$         & $\times$               & 18.2         & 16.4        & 18.2      & 19.3     \\ 
		$\times$             & $\checkmark$           & 21.0         & 18.7        & 20.6      & 21.9 \\
		$\checkmark$         & $\checkmark$           & \textbf{24.4}         & \textbf{22.5}        & \textbf{24.8}      & \textbf{26.4}        \\ \bottomrule
	\end{tabular}}

	\label{tab:ablation_EA2}
\end{table}

\begin{table}[]
	\caption{\textbf{Ablation study on CRM and ERM of BR}. CRM and ERM denote Category Refining Momentum and Category Refining Momentum, respectively. The results are acquired under the Transformer (FGPL-A) model.}
	\centering
	\resizebox{0.95\linewidth}{!}{
	\begin{tabular}{cc|cccc}
	\toprule 
		\multicolumn{2}{c}{BR}     & \multicolumn{4}{c}{Predicate Classification (PredCls)} \\ \hline
		CRM  & ERM   & mR@100/R@100   & DP@5  & DP@10  & DP@20  \\ \hline
		$\times$             & $\times$               & 41.0/50.1         & 38.0        & 40.7      & 42.5      \\
		$\times$             & $\checkmark$           & 42.0/50.9         & 38.5        & 41.1      & 42.8 \\
		$\checkmark$         & $\times$               & 41.9/53.3         & 38.4        & 41.0      & 42.7     \\ 
		$\checkmark$         & $\checkmark$           & \textbf{42.4/53.4}         & \textbf{38.6}        & \textbf{41.1}      & \textbf{42.9}        \\ \bottomrule
	\end{tabular}}

	\label{tab:ablation_BR}
\end{table}

\begin{table}[]
	\caption{\textbf{Ablation study on hard-to-distinguish threshold $\xi$ in CDL.} All results are obtained under the Transformer (CDL) model.}
	\centering
		\resizebox{\linewidth}{!}{
	\begin{tabular}{c|cccc}
	\toprule
	    \multirow{2}{*}{$\xi$}     & \multicolumn{4}{c}{Predicate Classification (PredCls)} \\ \cline{2-5}
		 & mR@20/R@20 & mR@50/R@50 & mR@100/R@100 &Mean \\ \hline
		0.7             & 22.7/43.0         & 30.6/54.0        & 33.9/57.7    &29.0/51.5      \\
		0.8        & 22.8/43.0         & 30.7/54.1        & 34.4/57.8       &29.3/51.6  \\
	\textbf{0.9}            & \textbf{23.0}/43.5         & \textbf{31.4/54.9}        &\textbf{35.4/58.5}    &\textbf{29.9/52.3}      \\
		1.0           & 22.6/43.6         & 30.3/54.7        & 33.8/58.2     &28.9/52.1     \\
		1.1         & 22.8/\textbf{43.7}         & 30.4/54.7        & 33.7/58.2      &28.9/52.2   \\ \bottomrule 
	\end{tabular}}

	\label{tab:ablation_EA1}
\end{table}

\noindent\textbf{Predicate Correlation (PC) \& Re-weighting Factor (RF) in CDL:} We explore the effectiveness of the Predicate Correlation (PC) and the Re-weighting Factor (RF) of CDL. To be specific, we discard PC by ignoring $\varphi_{i,j}  >\xi$ and $\varphi_{i,j} \le\xi$ in Eq.~\ref{predicate_correaltion}. Besides, we discard RF by setting Re-weighting Factor $w_{i,j}$ as 1 for all predicate pairs $i$  and $j$ in Eq.~\ref{balanced_softmax}. The results are shown in Tab.~\ref{tab:ablationCA_1}. It is worth noting that CDL (RF) leads to notable progress on mR@100 and DP@K, which demonstrates the effectiveness of RF. Furthermore, CDL outperforms the baseline with a considerable margin after being integrated with PC. We believe adjusting the re-weighting process based on PC, CDL improves the discriminatory power among hard-to-distinguish predicates while maintaining the original discriminating capability among recognizable ones.

\noindent\textbf{Predicate Correlation (PC) \& Balancing Factor (BF) in EDL:}
To validate the superiority for each component of EDL, i.e., Predicate Correlation (PC) and Balancing Factor (BF), we experiment with the following four settings: 1) Transformer with EDL (without PC and BF). 2) Transformer with EDL (without BF), i.e., removing the balancing factor $\frac{n_{j}}{n_{i}}$ in Eq.~\ref{eq:EDL}. 3) Transformer with EDL (without PC), i.e., setting $\mathcal{V}_i$ in Eq.~\ref{eq:EDL} as a set containing all predicate categories. 4) Transformer with EDL (with PC and BF). The experimental results are shown in Tab.~\ref{tab:ablation_EA2}. Without Predicate Correlation (PC), we observe a decrease on mR@100 ($24.4\%$ \textit{vs.}~$21.0\%$) and DP@K (e.g., DP@10: $24.8\%$ \textit{vs.}~$20.6\%$). It verifies the usefulness of PC for improving discriminatory capability. We attribute the behavior to the fact that EDL (PC) explores the underlying context information within each entity and adjusts the discriminating process based on gradually obtained discriminatory capability. Additionally, it can be observed that trained without BF, there is a reduction on mR@100 ($24.4\%$ \textit{vs.}~$18.2\%$) and DP@K (e.g., DP@10: $24.8\%$ \textit{vs.}~$18.2\%$), demonstrating the efficacy of BF for the enhanced discriminatory ability for hard-to-distinguish predicates. We think this may be caused by alleviating over-suppression to tail classes, which leads an efficient discriminating process among classes with different frequencies. Finally, discarding both PC and BF, we observe a more considerable margin of reduction on mR@50 and DP@K, demonstrating the strengths of both PC and BF.

\noindent\textbf{Batch-Refinement (BR) regime in PL-A:}
To prove that our proposed Batch-Refinement (BR) regime (i.e., Category Refining Momentum (CRM) and Entity Refining Momentum (ERM)) strengthens model's discriminatory power, we implement the following four settings in Tab.~\ref{tab:ablation_BR}: 1) BR without ERM and CRM, i.e., keeping ${\rm{s}_{i,j}^t=\rm{s}_{i,j}}$ throughout the
training procedure in Eq.~\ref{Eq:BR}. 2) BR with only ERM, i.e., discarding CRM ${\rm{\tilde s}}_{i,j}^t$ in Eq.~\ref{Eq:BR}. 3) BR with only CRM, i.e., ignoring ERM ${\rm{\hat s}}_{i,j}^t$ in Eq.~\ref{Eq:BR}. 4) BR with both CRM and ERM. Compared with results in the first line (BR without ERM and CRM) of Tab.~\ref{tab:ablation_BR}, Transformer (FGPL-A) with ERM or CRM in the second/third line achieves enhancements on both DP@K ($K=5, 10, 20$) and mR/R@100 metrics. Then, it performs the best after being integrated with both ERM and CRM. 
We owe such gains to the progressively refined predicate correlation, which adaptively regularizes model's learning procedure in keeping with its dynamic learning pace.   


\noindent\textbf{Hard-to-distinguish threshold $\xi$ in CDL:}
To investigate how the hard-to-distinguish threshold $\xi$ influences the discriminating process among predicates, we conduct experiments with different $\xi$. The quantitative results are shown in Tab.~\ref{tab:ablation_EA1}. 
The results illustrates that increasing $\xi$ makes the model gradually focus on discriminating among hard-to-distinguish predicates while preserving the learned discriminatory power of distinguishable ones. 
Particularly, it achieves the best on both mR@K and R@K with $\xi=0.9$ (i.e., Mean: $29.9\%/52.3\%$). 
After that, the continuous increase on $\xi$ (i.e., larger than $0.9$,) weakens the classifier's overall discriminatory power among predicates. 
Ultimately, we empirically figure out the best value for $\xi$ as $0.9$.

\noindent\textbf{Hyper-parameter $\tau$ in CRM:}
In this section, we attempt to explore the impacts of different $\tau$ (i.e., the hyper-parameter in CRM) on model's performance. 
Generally, as a trade-off coefficient, neither a small nor a large $\tau$ can balance the refining process of predicate correlations between the historical and the coming-batch statistics.
From Tab.~\ref{tab:ablation_gamma}, we observe that model's performance progressively improves with the increase of $\tau$, performing its best with $\tau=0.99$ (i.e., Mean: $36.3\%/47.3\%$), and then decreases with larger $\tau$.
Thus, we empirically set $\tau$ as $0.99$ in the CRM.

\begin{table}[]
	\caption{\textbf{Ablation Study on hyper-parameter $\tau$ of CRM.} All results are obtained under the Transformer (FGPL-A) model.}
	\centering
		\resizebox{\linewidth}{!}{
	\begin{tabular}{c|cccc}
	\toprule
	    \multirow{2}{*}{$\tau$}     & \multicolumn{4}{c}{Predicate Classification (PredCls)} \\ \cline{2-5}
	  & mR@20/R@20  & mR@50/R@50  & mR@100/R@100 & Mean  \\ \hline
		0.9        & 28.0/36.6     & 37.0/47.4    & 41.7/51.5      & 35.5/45.1    \\
		\textbf{0.99}             & 28.4/\textbf{38.9}     & \textbf{38.0/49.7}     & \textbf{42.4/53.4}   & \textbf{36.3/47.3} \\
		0.999        & \textbf{28.6}/38.9      & 37.7/49.7   & 42.0/53.4   &  36.1/47.3    \\\bottomrule 
	\end{tabular}}

	\label{tab:ablation_gamma}
\end{table}
\begin{table}[t]
	\caption{\textbf{Sentence-to-Graph Retrieval.} We evaluate the practicability of scene graphs generated by benchmark models integrated with FGPL-A on Sentence-to-Graph Retrieval task of the VG-SGG dataset.}
	\centering
		\resizebox{0.95\linewidth}{!}{
            \begin{tabular}{lcccccc}
	            \toprule	
            \multirow{2}{*}{Method} & \multicolumn{3}{c}{Gallery 1000}& \multicolumn{3}{c}{Gallery 5000} \\\cline{2-7}
                                      & R@20         & R@50         & R@100   & R@20         & R@50         & R@100     \\ \hline
            Transformer               & 14.4          & 26.4        & 35.9    & 3.7          & 8.0        & 14.4           \\
            ~-FGPL-A         &\textbf{26.0}         & \textbf{38.1}       &\textbf{51.8}      &\textbf{8.4}         & \textbf{16.2}       &\textbf{25.1}      \\ \hline
   
            Motif         & 15.7          & 29.8          & 45.9         & 4.1          & 8.5          & 15.6             \\
            ~-FGPL-A         &\textbf{22.9}         & \textbf{37.4}     &\textbf{52.1}         &\textbf{7.2}  &\textbf{14.9}   &\textbf{23.7}\\ \hline
            
            VCTree         & 15.1           & 27.9        & 39.0      & 4.5           & 8.2        & 15.0                 \\
            ~-FGPL-A         &\textbf{26.5}         & \textbf{41.4}     &\textbf{54.9}         &\textbf{8.4}         & \textbf{16.3}     &\textbf{25.3}\\
           \bottomrule
\end{tabular}}

	\label{tab.Sentence-to-Graph}
\end{table}

\begin{table}[t]
	\caption{\textbf{Image Captioning.} We evaluate the practicability of scene graphs generated by Transformer-SGG and Transformer-SGG (FGPL-A) on Image Captioning task of the VG-SGG dataset.}	\centering
		\resizebox{1.0\linewidth}{!}{
            \begin{tabular}{lcccc}
	            \toprule	
             Method             &Bleu-4 $\uparrow$   & Meteor $\uparrow$        & Cider $\uparrow$   & Spice $\uparrow$ \\ \hline
            Baseline               & 35.3          & 27.6     & 111.8   & 20.6       \\
            -Transformer     & 35.3          & 27.6     & 111.9   & 20.7 \\
            -Transformer(FGPL-A) & \textbf{35.3}     & \textbf{27.8}     & \textbf{112.2}    & \textbf{21.1}   \\
          \bottomrule
\end{tabular}}

	\label{tab.caption}
\end{table}

\begin{figure*}[t]
  \centering
\includegraphics[width=1\textwidth]{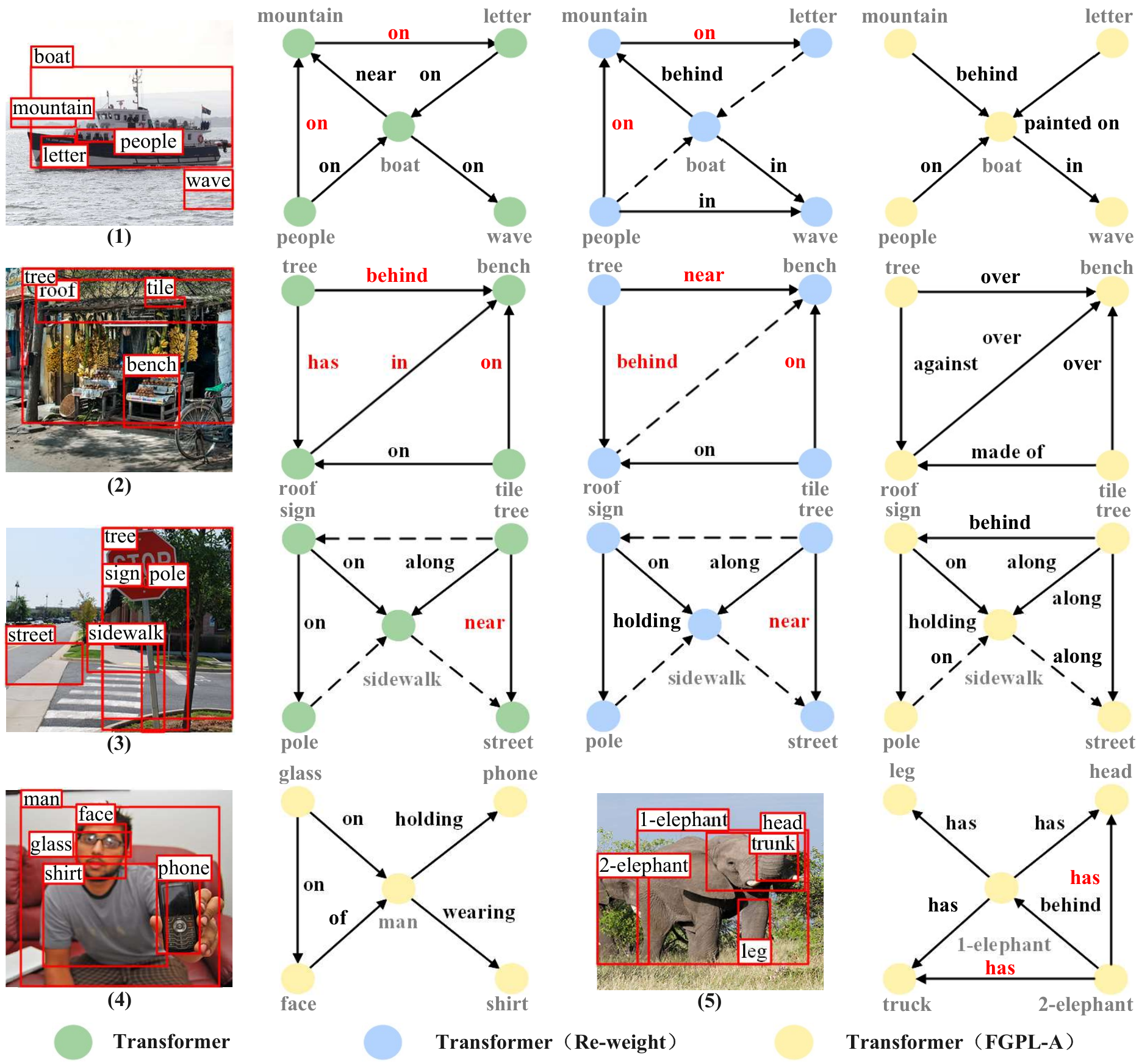}
\caption{\textbf{Scene graphs generated by Transformer, Transformer (Re-weight), and Transformer (FGPL-A) under the PredCls task.} Scene graphs generated by Transformer (FGPL-A) are more fine-grained than the vanilla Transformer, Transformer(Re-weight). For clarity, we omit some relationships.}
\label{fig:visual}
\label{visual}
\end{figure*}

\subsection{Practicability Analysis of FGPL-A}
To verify the practicability of fine-grained predicates within scene graphs generated by our FGPL-A, we conduct experiments on both Sentence-to-Graph Retrieval and Image Captioning tasks on the Visual Genome (VG) dataset.

\noindent\textbf{Sentence-to-Graph Retrieval:}
To affirm that scene graphs generated by our FGPL-A precisely describe the contents of images, we conduct the Sentence-to-Graph Retrieval task on the VG dataset, shown in Tab.~\ref{tab.Sentence-to-Graph}.
Specifically, scene graphs generated by benchmark models (i.e., Transformer, VCTree, and Motif) and our FGPL-A (e.g., Transformer (FGPL-A), VCTree (FGPL-A), Motif (FGPL-A)) are encoded to retrieve the corresponding textual scene graphs of image captions. 
Compared with benchmark models, we observe improvements on R@100 under Gallery 1000 from benchmarks trained with our FGPL-A learning strategies (e.g., Transformer (FGPL-A) ($51.8\%$ \textit{vs.}~$35.9\%$), VCTree (FGPL-A) ($52.1\%$ \textit{vs.}~$45.9\%$), Motif (FGPL-A) ($54.9\%$ \textit{vs.}~$39.0\%$)).
It validates FGPL-A's practicability in generating fine-grained scene graphs, which dramatically enriches the Sentence-to-Graph Retrieval task with precise scenario information.

\noindent\textbf{Image Captioning:}
Additionally, we evaluate the practicability of fine-grained scene graphs generated by FGPL-A utilizing the Image Captioning task on the VG dataset in Tab.~\ref{tab.caption}.
Practically, we implement three settings: 1) Baseline (Image Captioning model in ~\cite{cap:benchmark}) without features of scene graphs. 2) Baseline with features of scene graphs generated by the benchmark SGG model, namely Transformer. 3) Baseline with features of scene graphs generated by our Transformer (FGPL-A).
Utilizing the features of scene graphs generated by Transformer, it slightly outperforms the Baseline on the Cider and Spice metrics, which demonstrates the effectiveness of scene graphs on the scene-understanding task.
Furthermore, when taking advantage of scene graphs generated by our SGG model, i.e., Transformer (FGPL-A) outperform the Baseline by a larger enhancement on almost all the evaluation metrics (i.e., Blue-4: $35.3\%$ \textit{vs.}~$35.3\%$, Meter: $27.8\%$ \textit{vs.}~$27.6\%$, Cider: $112.2\%$ \textit{vs.}~$111.8\%$, Spice: $21.1\%$ \textit{vs.}~$20.6\%$).
It powerfully validates the practicability of fine-grained predicates generated by our FGPL-A.
Intuitively, since the FGPL-A regularizes SGG models with the comprehensive understanding on predicate correlations concerning contextual information, the generated scene graphs are prone to faithfully describe the contents of scenarios, benefiting scene-understanding tasks with rich semantics information.

\subsection{Visualization Results}
Finally, we testify the hypothesis that our proposed FGPL-A is capable of generating fine-grained predicates.
To the end, we make comparisons among scene graphs generated by Transformer, Transformer (Re-weight), and Transformer (FGPL-A) with the same input images from the VG-SGG dataset in Fig.~\ref{visual}. 

We observe that Transformer (FGPL-A) is able to generate more fine-grained relationships between objects than Transformer and Transformer (Re-weight), such as ``letter-\underline{painted on}-boat'' rather than ``letter-on-boat'' in Fig.~\ref{fig:visual} (1) and ``roof-over-bench'' instead of ``roof-in-bench'' in Fig.~\ref{fig:visual} (2).
From Fig.~\ref{fig:visual} (3), both Transformer and Transformer (Re-weight) fail to handle hard-to-distinguish predicates, i.e., ``near'' and ``along'', for the relationship between ``tree'' and ``street''.
In contrast, our Transformer (FGPL-A) successfully figures out the fine-grained predicate as ``along'', which is faithful to the scenario contexts. It powerfully demonstrates FGPL-A's effectiveness on generating fine-grained predicates for scene graphs.
Intuitively, comprehensively exploring the context information, FGPL-A can differentiate among hard-to-distinguish predicates.

Additionally, to further verify the effectiveness of FGPL-A in various scenarios, we visualize more scene graphs generated from Transformer (FGPL-A) shown in Fig.~\ref{fig:visual} (4) and Fig.~\ref{fig:visual} (5).
We observe that, in Fig.~\ref{fig:visual} (4), Transformer (FGPL-A) can precisely describe the interaction between ``man'' and ``phone'' as ``holding''. 
Similarly, in Fig.~\ref{fig:visual} (5), Transformer (FGPL-A) accurately infers the spatial relationship as ``behind'' between ``1-elephant'' and ``2-elephant''.
However, there are some unreasonable inference, e.g., ``\underline{2-elephant}-has-head'' and ``\underline{2-elephant}-has-truck'' shown in Fig.~\ref{fig:visual} (5), which are structural-inconsistent with ``\underline{1-elephant}-has-head'' and ``\underline{1-elephant}-has-truck''. 
Only considering the triplet-level context information, our method independently treats each ``subject-predicate-object'' during the inference process, while ignoring the structural information in the output space of generated scene graphs. Thus, how to generate structural-consistent scene graphs is still a challenging problem to be discussed in the future.


\section{Conclusion}
\label{sec:conclusions}
In this work, we first propose the fine-grained predicates problem, a new perspective of cases that hamper current SGG models. 
The address the problem, we propose a plug-and-play Adaptive Fine-Grained Predicates Learning (FGPL-A) framework for scene graph generation, which contributes to discriminating among hard-to-distinguish predicates with fine-grained ones.
In practice, we devise an Adaptive Predicate Lattice (PL-A) to help understand ubiquitous predicates correlation involving scenarios in SGG datasets and model's dynamic learning status at each learning pace. 
Based on the PL-A, we further develop an Adaptive Category Discriminating Loss (CDL-A) and an Adaptive Entity Discriminating Loss (EDL-A), which both help SGG models differentiate among hard-to-distinguish predicates while maintaining learned discriminatory power over recognizable ones throughout training.
Finally, comprehensive experiments show that our FGPL-A can differentiate among hard-to-distinguish predicates, benefits SGG models with a balanced discriminating learning procedure, and enriches downstream tasks with precise semantic information. 

\ifCLASSOPTIONcaptionsoff
  \newpage
\fi


\bibliographystyle{IEEEtran}
\bibliography{egbib}
	\begin{IEEEbiography}
		[{\includegraphics[width=1in,height=1.25in]{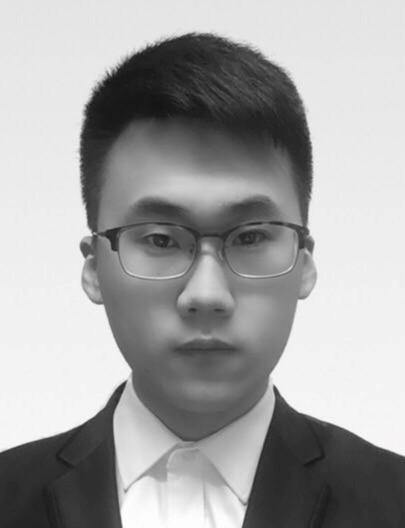}}]{Xinyu Lyu}
		is a PH.D. student in the School of Computer Science and Engineering, University of Electronic Science and Technology of China (UESTC). He is currently working on image understanding and scene graph generation.
	\end{IEEEbiography}
	
	\begin{IEEEbiography}
		[{\includegraphics[width=1in,height=1.25in]{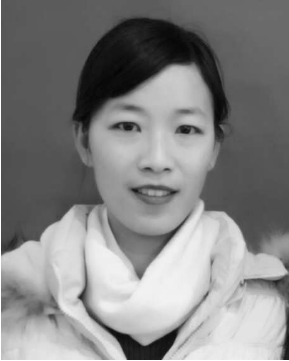}}]{Lianli Gao}
		(Member, IEEE)
		received the Ph.D. degree in information technology from The University of Queensland (UQ), Brisbane, QLD, Australia, in 2015. She is currently a Professor with the School of Computer Science and Engineering, University of Electronic Science and Technology of China (UESTC), Chengdu, China. She is focusing on integrating natural language for visual content understanding. Dr. Gao was the winner of the IEEE Trans. on Multimedia 2020 Prize Paper Award, the Best Student Paper Award in the Australian Database Conference, Australia, in 2017, the IEEE TCMC Rising Star Award in 2020, and the ALIBABA Academic Young Fellow.
	\end{IEEEbiography} 
	
	\begin{IEEEbiography}
		[{\includegraphics[width=0.95in,height=1.25in]{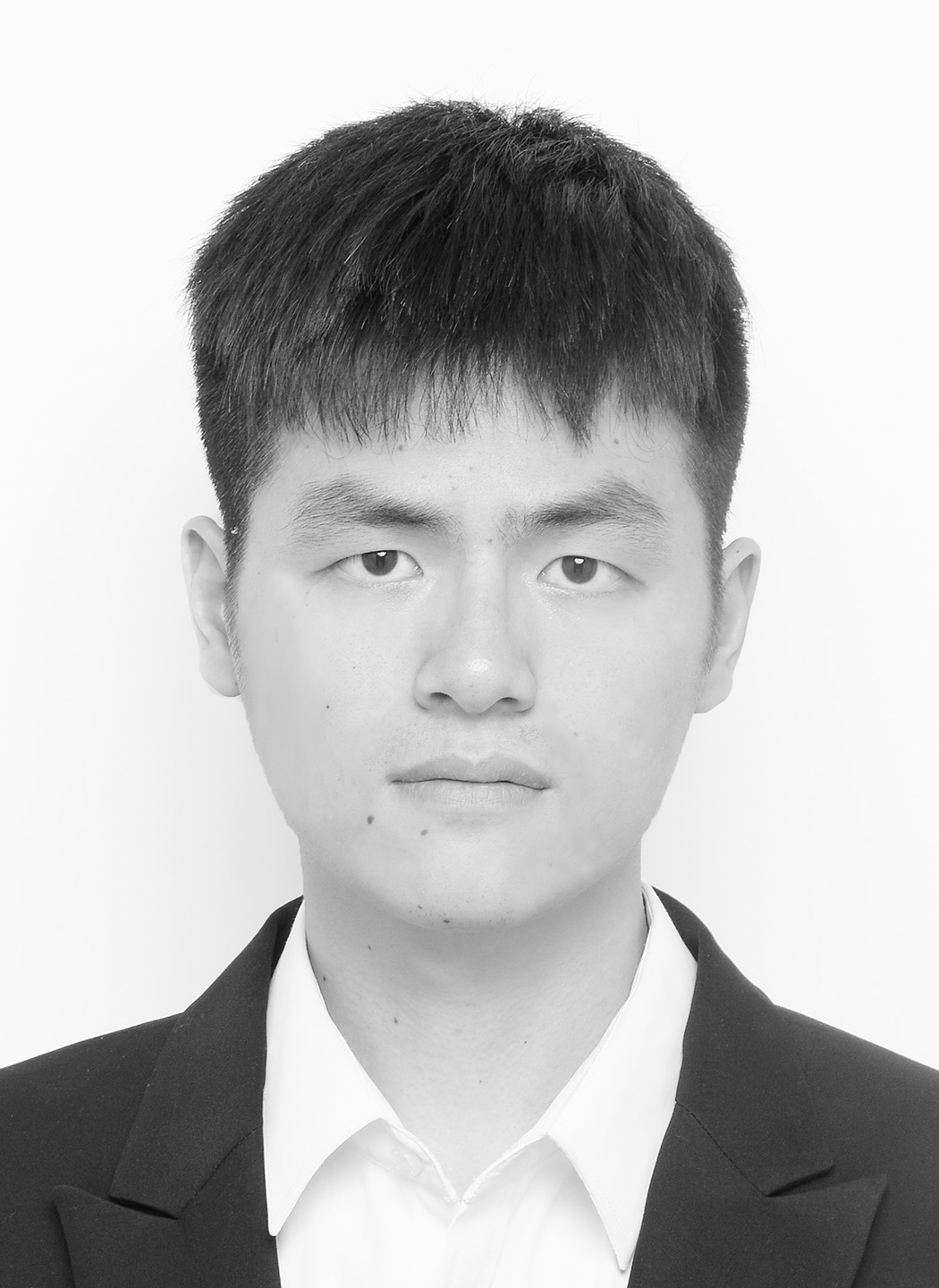}}]{Pengpeng Zeng} is a PhD candidate student in School of Computer Science and Engineering, UESTC. He is conducting research in visual understanding.
	\end{IEEEbiography}
	
	\begin{IEEEbiography}
		[{\includegraphics[width=1in,height=1.25in]{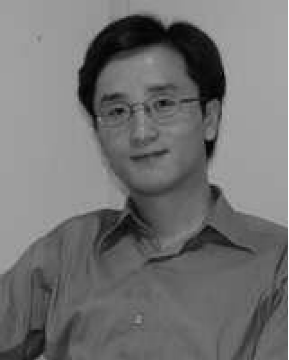}}]{Heng Tao Shen}(Fellow, IEEE)
		Heng Tao Shen is the Dean of School of Computer Science and Engineering, the Executive Dean of AI Research Institute at University of Electronic Science and Technology of China (UESTC). He obtained his BSc with 1st class Honours and PhD from Department of Computer Science, National University of Singapore in 2000 and 2004 respectively. His research interests mainly include Multimedia Search, Computer Vision, Artificial Intelligence, and Big Data Management. He is/was an Associate Editor of ACM Transactions of Data Science, IEEE Transactions on Image Processing, IEEE Transactions on Multimedia, IEEE Transactions on Knowledge and Data Engineering, and Pattern Recognition. He is a Member of Academia Europaea, Fellow of ACM, IEEE and OSA.  
	\end{IEEEbiography}
	
	\begin{IEEEbiography}
		[{\includegraphics[width=0.95in,height=1.25in]{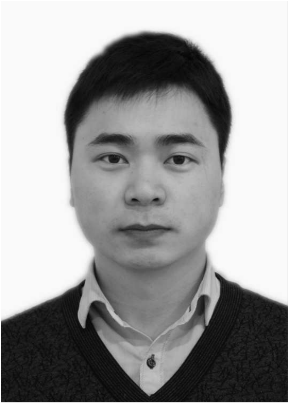}}]{Jingkuan Song} (Senior Member, IEEE) is currently a Professor with the University of Electronic Science and Technology of China (UESTC), Chengdu, China. His research interests include large-scale multimedia retrieval, image/video segmentation and image/video understanding using hashing, graph learning, and deep learning techniques. Dr. Song has been an AC/SPC/PC Member of IEEE Conference on Computer Vision and Pattern Recognition for the term 2018–2021, and so on. He was the winner of the Best Paper Award in International Conference on Pattern Recognition, Mexico, in 2016, the Best Student Paper Award in Australian Database Conference, Australia, in 2017, and the Best Paper Honorable Mention Award, Japan, in 2017.
	\end{IEEEbiography}
	
	
\end{document}